\documentclass[11pt,a4paper]{article}

\usepackage[margin=1in]{geometry}
\usepackage{setspace}
\usepackage{amsmath, amssymb, amsfonts}
\usepackage{physics}

\usepackage{graphicx}
\usepackage{subcaption}
\usepackage{array}
\usepackage{stackengine}

\usepackage{natbib} 

\usepackage{authblk}

\usepackage{fancyhdr}
\usepackage{xcolor}
\PassOptionsToPackage{hyphens}{url} 
\usepackage[colorlinks=true, linkcolor=blue, citecolor=blue, urlcolor=blue]{hyperref}

\title{\textbf{A Green-Integral--Constrained Neural Solver with Stochastic Physics-Informed Regularization}}
\date{} 

\author[1*]{Mohammad Mahdi Abedi}
\author[1,2,3]{David Pardo}
\author[4]{Tariq Alkhalifah}

\affil[1]{University of the Basque Country (UPV/EHU), Department of Mathematics, Spain}
\affil[2]{Basque Center for Applied Mathematics (BCAM), Bilbao, Spain}
\affil[3]{Ikerbasque, Basque Foundation for Science, Bilbao, Spain}
\affil[4]{King Abdullah University of Science and Technology (KAUST), Thuwal, Saudi Arabia}
\affil[*]{Corresponding author: \href{mailto:mohammadmahdi.abedi@ehu.eus}{mohammadmahdi.abedi@ehu.eus}}

\begin{document}
		\graphicspath{
		{"Figures/"}}
	\maketitle
	\thispagestyle{empty} 
	
\begin{abstract}
	Standard physics-informed neural networks (PINNs) struggle to simulate highly oscillatory Helmholtz solutions in heterogeneous media because pointwise minimization of second-order PDE residuals is computationally expensive, biased toward smooth solutions, and requires artificial absorbing boundary layers to restrict the solution. To overcome these challenges, we introduce a Green--Integral (GI) neural solver for the acoustic Helmholtz equation. It departs from the PDE-residual-based formulation by enforcing wave physics through an integral representation that imposes a nonlocal constraint. Oscillatory behavior and outgoing radiation are encoded directly through the integral kernel, eliminating second-order spatial derivatives and enforcing physical solutions without additional boundary layers. Theoretically, optimizing this GI loss via a neural network acts as a spectrally tuned preconditioned iteration, enabling convergence in heterogeneous media where the classical Born series diverges. By exploiting FFT-based convolution to accelerate the GI loss evaluation, our approach substantially reduces GPU memory usage and training time. However, this efficiency relies on a fixed regular grid, which can limit local resolution. To improve local accuracy in strong scattering regions, we also propose a hybrid GI+PDE loss, enforcing a lightweight Helmholtz residual at a small number of nonuniformly sampled collocation points. We evaluate our method on seismic benchmark models characterized by structural contrasts and subwavelength heterogeneity at frequencies up to 20~Hz. GI-based training consistently outperforms PDE-based PINNs, reducing computational cost by over a factor of ten. In models with localized scattering, the hybrid loss yields the most accurate reconstructions, providing a stable, efficient, and physically grounded alternative.
	
	\vspace{0.5cm} 
\end{abstract}

\section{Introduction}

Wave propagation in heterogeneous media is governed by complex interference and multiple-scattering phenomena that challenge both numerical accuracy and computational efficiency across a wide range of applications, including seismic imaging, medical ultrasound, and acoustics. In the frequency domain, wavefield simulation is commonly formulated in terms of the acoustic Helmholtz equation, whose solution describes the steady-state response of the medium to harmonic sources.  
Classical numerical methods for solving the Helmholtz equation, such as finite-difference (FD), finite-element (FE), and spectral-element methods, are well established and highly accurate, but they suffer from substantial computational and memory costs at high frequencies due to the need for fine spatial discretization and the resulting large linear systems that should be solved.

Neural-network-based approaches have emerged as a more flexible paradigm for solving partial differential equations (PDEs).  Among them, Physics-Informed Neural Networks (PINNs) \cite{raissi2019} have attracted significant attention due to their ability to incorporate governing physics of the problem directly into the training process through automatic differentiation. By minimizing the residual of the PDE at a set of collocation points, PINNs offer a mesh-free framework that approximates wavefields as differentiable functions, avoiding the need for traditional grid-based discretization of differential operators.

During training, a PINN searches for a solution by minimizing the PDE residual at a finite set of collocation points, effectively exploring the solution space through gradient-based optimization guided by local differential constraints. However, local spatial derivatives often fail to capture complex global behaviors, prompting the development of nonlocal PINN architectures \citep{haghighat2021nonlocal}. For  highly oscillatory wave equations, the local enforcement leaves a large space of admissible solutions that approximately satisfy the PDE but differ significantly in their global physical behavior. As a result, the optimization landscape associated with the Helmholtz operator is highly nonconvex, leading to unstable training dynamics characterized by slow convergence and strong sensitivity to initialization. These difficulties become more pronounced when the number of collocation points is limited, allowing the network to exploit unsampled regions of the domain, or when strong scattering produces complex interference patterns that cannot be adequately constrained. These challenges have motivated different research directions aimed at improving the effectiveness of the training data, the properties of the network, and the optimization strategies.

Substantial effort has been devoted to improving both the representational capacity and the optimization behavior of neural networks for highly oscillatory PDEs. On the representation side, enriched input parameterizations such as sinusoidal positional encoding and multiresolution hash encoding have been introduced to enable neural networks to better capture high-frequency content \citep{Tancik2020,muller2022, huang2024efficient}. 
Related approaches incorporate oscillatory inductive biases into the network architecture, for example, through Fourier-feature mappings, Gabor-type basis functions, or adaptive sinusoidal representation networks which embed frequency information and have shown improved expressiveness for wave phenomena \citep{sitzmann2020implicit, Huang2023, wu2024pinn,abedi2025gabor,zhang2026adaptive}. Representing solutions in the wavelet domain has also been proposed to handle oscillations while bypassing the high computational cost of automatic differentiation \citep{pandey2026efficient}. Besides, hybrid training strategies \citep{cyr2020robust,abedi2025least}, adaptive activation functions \citep{ jagtap2020locally,song2022versatile,wu2024pinn}, domain decomposition methods \citep{ kharazmi2021hp}, curriculum-based learning \citep{ krishnapriyan2021characterizing}, modified loss balancing schemes \citep{ wang2021understanding}, and other state-of-the-art methods have been proposed to improve robustness and convergence.

The number and distribution of collocation points also affect the performance of PINNs. Various sampling methods, such as uniform, random, and dynamic adaptive sampling, are used in PINNs training \citep{ luo2025physics,omella2024r}. More recently, residual-based adaptive sampling strategies are proposed to locally increase collocation density in regions where the PDE residual is large, based on current loss indicators \citep{lu2021deepxde,nabian2021efficient,wu2023comprehensive}. However, for wave propagation problems governed by the Helmholtz equation, the source of a large residual at a given point can be nonlocal: a perturbation or mismatch far from that point may produce high residuals through constructive interference, reflections, or diffraction effects. Therefore, when errors in wavefields have long-range, nonlocal causes, treating them as local phenomena can mislead the adaptive sampling process.

The reliance of PINNs on local differential constraints continues to shape their behavior in subtle but important ways. 
 Several studies have reported that training PINNs using only the PDE residual, without explicitly enforcing absorbing boundary conditions, does not produce classical boundary reflections (e.g., \cite{alkhalifah2021,Rasht-Behesht2022,wu2023,song2021,ding2025physics}). 
This behavior can be attributed to the pointwise evaluation of the differential residual and to the implicit bias of neural networks, which favors learning smooth continuations of interior patterns. Nevertheless, in the absence of an explicit restriction on the admissible solution space, PDE-only formulations remain vulnerable to convergence toward spurious, trivial, or otherwise nonphysical wavefields, particularly in strongly scattering media \citep{ding2025physics}.
A classical remedy is to introduce perfectly matched layers (PMLs) combined with Dirichlet boundary conditions \citep{abedi2025gabor}, at the cost of increased computational and memory requirements due to domain padding. Yet, we observe that in strongly scattering media, even with PMLs properly enforced, collocation based optimization remains  vulnerable to converging toward spurious wavefields.

In this work, we adopt a different perspective by introducing a global physical constraint derived from the Lippmann--Schwinger integral equation. We propose a Green--Integral (GI) neural network solver for the acoustic Helmholtz equation that restricts the admissible solution space by enforcing consistency between the neural-network-predicted scattered field and its GI reconstruction. Unlike the PDE residual, which imposes local differential constraints and requires repeated evaluation of high-order derivatives by automatic differentiation, the GI term enforces global consistency of the wavefield, couples the solution at each spatial location to all points in the domain, and eliminates the need for costly derivative computations.
As a result, it inherently satisfies the radiation condition and enforces outgoing-wave behavior without requiring sophisticated absorbing boundary conditions or PML extensions.

A practical limitation of our GI formulation is that its efficient evaluation requires the scattered field to be represented on a uniform regular grid. On such a grid, compact or strong scatterers may be under-resolved, leading to subgrid-scale effects that are not fully captured by the integral constraint alone. This motivates the inclusion of a complementary PDE-based loss evaluated at collocation points sampled nonuniformly based on the scattering potential, providing targeted local correction without altering the global integral formulation. By concentrating PDE-based local corrections in regions with higher scattering potential, the neural network is encouraged to focus its refinement on the most complex areas.

\textbf{Contributions}: The main contributions of this work are summarized as follows:
\begin{itemize}
	\item A Green--Integral neural network solver for the acoustic Helmholtz equation that enforces physical consistency through the Lippmann--Schwinger formulation, eliminating the need for  boundary conditions and spatial derivatives.
	
	\item An FFT-accelerated implementation of the GI loss that exploits the convolutional structure of the integral formulation, enabling scalable and memory-efficient training on dense grids.
	
	\item A hybrid GI+PDE training strategy that augments the global integral constraint with local PDE corrections to improve accuracy in regions of strong scattering.
	
	\item A theoretical interpretation of the proposed training procedure that connects neural network optimization of the GI loss to classical iterative solvers for the Lippmann--Schwinger equation, revealing the network architecture as an implicit learned preconditioner.
	
	\item A demonstration of substantial improvements in training stability, accuracy, and computational efficiency compared to conventional PDE-residual-based PINN formulations, using multiple benchmark velocity models.
\end{itemize}

The remainder of this paper is organized as follows. 
Section~\ref{sec:method} presents the mathematical formulation of the scattered Helmholtz equation and the associated Green--Integral representation. 
Section~\ref{sec:Methodology} describes the numerical methodology and implementation details. Section~\ref{sec:connection_to_classical} analyzes the connection between the proposed approach and classical iterative solvers.  Section~\ref{sec:results} presents numerical experiments and comparisons with conventional PINN formulations. 
Section~\ref{sec:discussion} discusses the physical interpretation of the results, identifies the regimes in which the method is most effective, and outlines its limitations.

\section{Problem Statement}
\label{sec:method}

We consider the frequency-domain acoustic wave propagation problem and describe both its differential and integral formulations for the scattered wavefield.

\subsection{Helmholtz Equation for scattered wavefield}

The time-harmonic wave equation governing acoustic wave propagation is given by
\begin{equation}
	\left( \nabla^2 + \omega^2 m(\mathbf{x}) \right) U(\mathbf{x}) = -f(\mathbf{x}),
	\label{eq:helmholtz}
\end{equation}
where \( U(\mathbf{x}) \) is the total complex-valued wavefield,  
\( \omega \) is the angular frequency,  
\( m(\mathbf{x}) = 1/v^2(\mathbf{x}) \) denotes the squared slowness (where $v$ is the velocity),  
and \( f(\mathbf{x}) \) represents the source term.

We decompose the medium into a constant background ($m_0$) and a perturbation component ($\delta m$), as follows:
\[
m(\mathbf{x}) = m_0(\mathbf{x}) + \delta m(\mathbf{x}),
\]
and express the total wavefield as the sum of a background ($U_0$) and scattered wavefield ($U_s$), 
\[
U(\mathbf{x}) = U_0(\mathbf{x}) + U_s(\mathbf{x}),
\]
where the background wavefield \( U_0 \) satisfies
\begin{equation}
	\left( \nabla^2 + \omega^2 m_0(\mathbf{x}) \right) U_0(\mathbf{x}) = -f(\mathbf{x}).
\end{equation}
Subtracting this equation from \eqref{eq:helmholtz} yields the scattered-field Helmholtz equation:
\begin{equation}
	\left( \nabla^2 + \omega^2 m(\mathbf{x}) \right) U_s(\mathbf{x}) = -\omega^2 \delta m(\mathbf{x}) U_0(\mathbf{x}).
	\label{eq:scattered}
\end{equation}

The scattered-field equation is used to avoid the point-source singularity in the solution \citep{alkhalifah2021}.  However, enforcing the scattered Helmholtz equation by pointwise minimization of the PDE residual does not uniquely determine the physical scattering solution and may lead to non-physical or overly smooth solutions that are locally consistent with the operator in the interior but violate global scattering behavior. The physically relevant scattered field is uniquely characterized by the nonlocal Sommerfeld radiation condition. While absorbing boundaries, such as perfectly matched layers (PMLs), are classically introduced to approximate this condition, their explicit enforcement in a neural network framework increases the optimization complexity and cost.

Importantly, in a collocation-based PINN framework, the loss landscape remains effectively non-unique even when a PML is utilized. Because the network does not sample the exact location of the source singularity (a Dirac delta function), minimizing the residual of \eqref{eq:scattered} at a finite set of collocation points admits the trivial total-field solution, $U_s = -U_0$. With or without PML, this spurious solution yields a zero PDE residual at all source-free collocation points, producing a non-physical $U_s$. 

To overcome this issue, we use an alternative formulation that directly encodes the nonlocal physics of wave propagation and the Sommerfeld radiation condition without requiring artificial boundary layers.

	\subsection{Green--Integral Formulation}
	
	The Green--Integral (GI) formulation expresses the scattered wavefield as a nonlocal function of the total field and medium perturbation.  
	Starting from the scattered-field Helmholtz equation \eqref{eq:scattered}, we apply Green's representation theorem using the Green's function of the background medium \(G_0(\mathbf{x}, \mathbf{y})\), defined as the solution of
	\begin{equation}
		\left( \nabla^2 + \omega^2 m_0(\mathbf{x}) \right) G_0(\mathbf{x}, \mathbf{y}) = -\delta(\mathbf{x} - \mathbf{y}),
	\end{equation}
where \(\delta(\cdot)\) is the Dirac delta function,
\(\mathbf{x}\) and \(\mathbf{y}\) denote the observation and integration locations, respectively, \(\omega\) is the angular frequency, \(m_0(\mathbf{x}) = 1/v_0^2(\mathbf{x})\), and \(v_0\) is the background velocity. Here, \(\mathbf{y}\) represents a scattering location within the medium.

	The resulting Lippmann--Schwinger integral equation for the scattered field is a GI representation of the scattered wavefield \citep{lippmann1950variational, colton2013integral}:
	\begin{equation}
		U_s(\mathbf{x}) = \omega^2 \int_{\Omega} G_0(\mathbf{x}, \mathbf{y})\,\delta m(\mathbf{y})\,U(\mathbf{y})\,d\mathbf{y},
		\label{eq:green_integral}
	\end{equation}
	where \(\delta m(\mathbf{y}) = m(\mathbf{y}) - m_0(\mathbf{y})\) denotes the squared-slowness perturbation, \(U(\mathbf{y}) = U_0(\mathbf{y}) + U_s(\mathbf{y})\) is the total field, \(\Omega\) is the computational domain, and $ G_0$ acts as a propagator.
	 This equation is \emph{exact} when the perturbation
	\(\delta m(\mathbf{y})\) is \emph{compactly supported}, i.e., nonzero only inside a bounded
	subdomain. 
	To approximate this condition in practice, the velocity model is extended by padding the computational
	domain and applying a smooth taper such that \(\delta m \rightarrow 0\) toward the boundaries.
	This ensures that the Green's function \(G_0\) correctly
	represents outgoing waves from a finite scattering region.

	Equation \eqref{eq:green_integral} establishes a \emph{global physical coupling}: the scattered field at any observation point \(\mathbf{x}\) depends on contributions from all scatterer locations \(\mathbf{y}\), weighted by the background Green's function.  
	It also provides a compact and physically constrained description of the scattering process that naturally incorporates radiation behavior.

	In a homogeneous background with constant velocity \(v_0\), the Green's function has analytical expressions in two- and three-dimensional media:
	\[
	G_0^{2D}(\mathbf{x}, \mathbf{y}) = \frac{i}{4} H_0^{(2)}( \frac{\omega}{v_0}\|\mathbf{x}-\mathbf{y}\|), 
	\quad
	G_0^{3D}(\mathbf{x}, \mathbf{y}) = \frac{e^{i\|\mathbf{x}-\mathbf{y}\|\omega / v_0}}{4\pi\|\mathbf{x}-\mathbf{y}\|},
	\]
	where \(H_0^{(2)}\) is the Hankel function of the second kind and $\|.\|$ is the Euclidean distance.

By replacing local second-order differential constraints of the scattered Helmholtz equation with a global integral relation involving the background Green's function, the GI formulation avoids explicit evaluation of the Laplacian operator while naturally incorporating radiation behavior and long-range wave interactions.

\section{Methodology}
\label{sec:Methodology}

Our objective is to learn the scattered wavefield \(U_s(\mathbf{x})\) using a neural network representation that satisfies the Green--Integral (GI) formulation \eqref{eq:green_integral} of the scattering problem.  

To evaluate the GI relation numerically, the computational domain $\Omega$ is discretized into $N_y$ scatterer points $\{\mathbf{y}_k\}_{k=1}^{N_y}$ forming a regular dense grid. The scattered field $U_s$ is evaluated at $N_x$ observation points $\{\mathbf{x}_j\}_{j=1}^{N_x}$. While the formulation allows these observation points to be placed arbitrarily, our implementation in the next subsection defines them to coincide with the scatterer grid.
The discrete approximation of the integral reads:
\begin{equation}
	\widehat{U}_s(\mathbf{x}_j) = 
	\omega^2 \sum_{k=1}^{N_y} 
	G_0(\mathbf{x}_j, \mathbf{y}_k)\,
	\delta m(\mathbf{y}_k)\,
	\big[ U_0(\mathbf{y}_k) + U_s(\mathbf{y}_k) \big]\,W_k,
	\label{eq:green_discrete}
\end{equation}
where \(W_k\) are the quadrature weights corresponding to the cell area (or volume) around each \(\mathbf{y}_k\).
The quantity \(\widehat{U}_s(\mathbf{x}_j)\) represents the GI \emph{reconstructed scattered field} from the current neural network prediction of \(U_s(\mathbf{y}_k)\).

Defining \(\mathbf{G}\in\mathbb{C}^{N_x\times N_y}\) as the background Green's function matrix,  
whose entries are given by \(G_{jk} = G_0(\mathbf{x}_j, \mathbf{y}_k)\), the discretized equation \eqref{eq:green_discrete} can be expressed as
\begin{equation}
	\widehat{\mathbf{u}}_{s} = \mathbf{G}\,\mathbf{d},
	\label{eq:matrix_Green--Integral}
\end{equation}
where the vector \(\mathbf{d} \in \mathbb{C}^{N_y}\) collects the local scattering sources
\begin{align*}
	d_k = \omega^2 \, \delta m(\mathbf{y}_k)\, W_k \,\Big(U_0(\mathbf{y}_k) + U_s(\mathbf{y}_k)\Big),
	\quad k=1,\dots,N_y.
\end{align*}

In this matrix form, \eqref{eq:matrix_Green--Integral} represents a dense linear mapping between the scattering sources \(\mathbf{d}\) and the reconstructed scattered field \(\widehat{\mathbf{u}}_s\).  
Although mathematically straightforward, direct evaluation scales as 
\(\mathcal{O}(N_x N_y)\) in both computation and memory.  

\subsection{Computation via FFT-Based Convolution}

When the background velocity \(v_0\) is constant, the Green's function \(G_0(\mathbf{x}, \mathbf{y})\) is proportional only to the Euclidean distance \(r = \|\mathbf{x}-\mathbf{y}\|\).  
In this case, the discrete Green's matrix \(\mathbf{G}\) exhibits a \emph{Toeplitz} (or block-Toeplitz in multiple dimensions) structure, meaning that each row is a shifted version of the first.  
This property allows the matrix-vector product in \eqref{eq:matrix_Green--Integral} to be expressed as a multi-dimensional convolution between the discrete Green's kernel ($G_0$) and the field-dependent quantity ($D$):
\begin{equation}
	\begin{split}
		\widehat{U}_s(\mathbf{x}) &= (G_0 * D)(\mathbf{x}), \\
		D(\mathbf{y}) &= \omega^2 \, \delta m(\mathbf{y}) \big[ U_0(\mathbf{y}) + U_s(\mathbf{y}) \big] W,
	\end{split}
\end{equation}
where \( * \) denotes convolution over the spatial coordinates.

For the Fast Fourier Transform (FFT)-based evaluation of the GI term, we restrict the GI loss to a fixed regular grid where the observation and integration points coincide.
That is, for the GI constraint, we evaluate the scattered field only at locations
\(\mathbf{x}_k = \mathbf{y}_k\), \(k=1,\dots,N_y\).
From this point onward, we denote both the observation and integration locations in the GI loss by \(\mathbf{y}\) for simplicity.
The PDE residual, when used, is evaluated independently at a separate set of collocation points \(\{\mathbf{x}_j\}_{j=1}^{N_x}\). 

The convolution can then be computed efficiently using FFT:
\begin{equation}
	\widehat{U}_s (\mathbf{y}) = \mathcal{F}^{-1}\!\left\{\,\mathcal{F}(G_0)\,\cdot\,\mathcal{F}(D)\,\right\},
	\label{eq:fft_convolution}
\end{equation}
where \(\mathcal{F}\) and \(\mathcal{F}^{-1}\) denote the forward and inverse Fourier transforms, and \(\cdot\) represents elementwise multiplication in the Fourier domain.  
FFT-based evaluation avoids the explicit formation of the dense matrix \(\mathbf{G}\),  requires only 
\(\mathcal{O}(N_y)\) memory 
and can be computed efficiently in 
\(\mathcal{O}(N_y \log N_y)\) time.

\subsection{Loss Function}

The GI equation provides the primary physical constraint by enforcing global consistency between the neural-network-predicted scattered field \(U_s(\mathbf{y}_j)\) and the reconstructed field \(\widehat{U}_s(\mathbf{y}_j)\) obtained from equation~\eqref{eq:fft_convolution}.  
The corresponding GI loss function is defined as
\begin{equation}
	\mathcal{L}_{\text{GI}} 
	= \frac{1}{N_y} \sum_{j=1}^{N_y}
	\left| 
	\widehat{U}_s(\mathbf{y}_j) - U_s(\mathbf{y}_j)
	\right|^2,
	\label{eq:GI_loss}
\end{equation}
where \(N_y\) is the number of integration (scatterer) grid points.

This loss penalizes violations of the GI relation~\eqref{eq:green_discrete}, thereby constraining the network to produce scattered wavefields that are globally consistent with the background Green's function and the distribution of scattering potential. It restricts the admissible solution space and suppresses trivial or spurious solutions such as $U_s = -U_0$.

A complementary local physical constraint can be imposed through the scattered Helmholtz equation~\eqref{eq:scattered}.  
The residual of this equation is evaluated at a set of collocation points \(\{\mathbf{x}_i\}_{i=1}^{N_x}\) to define the PDE loss:

\begin{equation}
 \mathcal{L}_{\text{PDE}} = \frac{1}{N_x} \sum_{i=1}^{N_x} \Big| \nabla^2 U_s(\mathbf{x}_i) + \omega^2 m(\mathbf{x}_i)\,U_s(\mathbf{x}_i) + \omega^2\,\delta m(\mathbf{x}_i) U_0(\mathbf{x}_i) \Big|^2%
	\label{eq:PDE_loss}
\end{equation}
The hybrid loss function combines the global GI loss and the local PDE residual loss:
\begin{equation}
	\mathcal{L_{\text{hybrid}} }
		= \mathcal{L}_{\text{GI}} 
		+ \lambda\,\mathcal{L}_{\text{PDE}} 
		\label{eq:total_loss}
	\end{equation}
	where \(\lambda\) is a scalar weight controlling the relative importance of the loss terms.

	The loss terms operate on different numerical scales.  The PDE loss \(\mathcal{L}_{\text{PDE}}\) measures the residual of the Helmholtz operator and thus depends on the spatial derivatives of the predicted wavefield, while the GI loss, \(\mathcal{L}_{\text{GI}}\), measures the squared error of the wavefield amplitudes (with the same scale of a data misfit loss).
 To reduce this discrepancy, all spatial coordinates are nondimensionalized with respect to one background wavelength, i.e.,
\begin{equation}
	\tilde{\mathbf{x}} = \frac{\omega\,\mathbf{x}}{2 \pi v_0}.
			\label{eq:normalization}
	\end{equation}
	
	By evaluating the GI term over a dense regular grid and the more expensive PDE term over a smaller set of randomly sampled collocation points, the training process can efficiently enforce both global and local physical constraints.  
	
\subsection{Implementation}
\label{sec:implementation}

The scattered wavefield ${U}_s(\mathbf{x})$ is represented by a fully connected multi-layer perceptron (MLP) with sinusoidal activation functions. To better represent oscillatory wavefields, we employ a sinusoidal encoder at the input layer that maps the spatial coordinates to periodic features. This encoding improves the network's ability to represent high-wavenumber components and reduces the neural network's spectral bias.

The background wavefield $U_0$ and the Green's function kernel $G_0$ depend only on the background medium and the frequency. Since these quantities remain fixed during training, they are precomputed once prior to optimization and reused throughout the training process. The evaluation of the GI term is then performed using FFT-based convolution, which efficiently computes the discrete convolution between $G_0$ and the scattering source term on a regular grid.

To avoid wrap-around artifacts inherent to FFT-based convolution, the Green's function kernel $G_0$ is constructed on a grid that is at least twice the size of the physical computational domain in each spatial dimension. 

\paragraph{Singularity of Green's function}
The two-dimensional Green's function exhibits a logarithmic singularity when an
observation point coincides with its source location.
While this singularity is integrable in the continuous formulation of
equation~\eqref{eq:green_integral}, its numerical evaluation on a discrete grid
requires special care.
In FFT-based implementations, the singular contribution corresponds to the
self-interaction of a grid cell with itself, i.e., the zero-offset term $G_0(\mathbf{0})$.
A simple and commonly used remedy is to set $G_0(\mathbf{0})=0$, which removes this nonphysical self-interaction, but introduces a first-order error that decays as the grid is refined. For improved accuracy, we calculate $G_0(\mathbf{0})$ by a cell-averaged Green's function derived in Appendix  \ref{app:self_term}. It provides an accurate approximation to the self-term of the integral that is compatible with our FFT-based implementation.

\paragraph{Sampling for the PDE loss} For the calculation of the PDE residual, before training begins, we generate a large pool of random collocation points, interpolate $v(\mathbf{x})$, and calculate $U_0(\mathbf{x})$ for these points. During training, at each training epoch, we randomly select $N_x$ points from the pool. 

When using the hybrid GI+PDE loss, we concentrate the PDE collocation points in regions of high scattering potential so that the PDE term locally complements the global GI term. We define the selection probability $P(\mathbf{x}_i)$ for each candidate collocation point
 $\mathbf{x}_i$ within the set of all available points $\mathcal{X}$ as:
\begin{equation}
	P(\mathbf{x}_i) = \frac{\mathcal{I}(\mathbf{x}_i)}{\sum_{\mathbf{x}_j \in \mathcal{X}} \mathcal{I}(\mathbf{x}_j)},
\end{equation}
where the importance function $\mathcal{I}(\mathbf{x})$ is defined based on the magnitude of the scattering potential $\delta m(\mathbf{x})$:
\begin{equation}
	\mathcal{I}(\mathbf{x}) = |\delta m(\mathbf{x})|^\alpha + \epsilon.
\end{equation}
In this formulation, $\alpha$ is a sharpening hyperparameter that controls the sampling density contrast, and $\epsilon$ is a small constant ensuring non-zero coverage in the homogeneous background. In our tests, we use $\alpha=1$ , and $\epsilon = 0.01 \, \delta m_{\text{max}}$.

To maintain computational efficiency, instead of performing probabilistic selection of collocation points at each epoch, we bias the spatial distribution of the pre-calculated pool to be more concentrated in areas with higher scattering potential, using the probabilistic selection of $N_{pool}$ points from a larger set of random points. During training, we then utilize a simple uniform random selection from this pool. This approach avoids the recurring $O(N_{pool})$ cost of probabilistic selection at each epoch while preserving the stochastic benefits of random collocation.

\section{Theoretical Relation to Classical Iterative Solvers}
\label{sec:connection_to_classical}

To provide a mathematical framework for analyzing the convergence behavior of the proposed Green--Integral (GI) neural solver, this section casts the optimization of the integral loss as an iterative scheme and compares it to the classical Born--Neumann series.

Starting from the discrete GI formulation \eqref{eq:green_discrete}, 
and using the matrix representation,  the scattered field can be written in compact form. In the case where the observation and scatterer grids coincide, the scattered field 
$\mathbf{u}_s \in \mathbb{C}^{N_y}$ should satisfy
\begin{equation}
	\mathbf{u}_s = \mathbf{G}\mathbf{M}\,(\mathbf{u}_0 + \mathbf{u}_s),
\end{equation}
where $\mathbf{G} \in \mathbb{C}^{N_y \times N_y}$ denotes the background Green's function matrix, and $\mathbf{M} \in \mathbb{C}^{N_y \times N_y}$ is a diagonal matrix with entries $M_{kk} = \omega^2 \delta m(\mathbf{y}_k) W$.

To facilitate the convergence analysis, we define the scattering operator
\[
\mathbf{A} = \mathbf{G}\mathbf{M} \in \mathbb{C}^{N_y \times N_y},
\]
and the incident source term
\[
\mathbf{b} = \mathbf{G}\mathbf{M}\mathbf{u}_0 \in \mathbb{C}^{N_y},
\]
corresponding to the first-order scattering (Born approximation). The Lippmann--Schwinger equation can then be written equivalently as the linear system
\begin{equation}
	(\mathbf{I} - \mathbf{A})\,\mathbf{u}_s = \mathbf{b},
	\label{eq:LiSch_linear_system}
\end{equation}
where $\mathbf{I}$ denotes the identity matrix. For an approximate solution $\mathbf{u}_s^{(n)}$, the residual associated with this system is defined as
\begin{equation}
	\mathbf{r}^{(n)} = (\mathbf{I}-\mathbf{A})\mathbf{u}_s^{(n)}-\mathbf{b}.
	\label{eq:residual_def}
\end{equation}
Different numerical solvers for equation \eqref{eq:LiSch_linear_system} can be interpreted in terms of how they use this residual to update the solution.

\subsection{The Born--Neumann Series}

The classical Born--Neumann series corresponds to solving \eqref{eq:LiSch_linear_system} via an explicit fixed-point (Picard) iteration. Starting from an initial guess $\mathbf{u}_s^{(0)}$, the update takes the form
\begin{equation}
	\mathbf{u}_s^{(n+1)} = \mathbf{A}\mathbf{u}_s^{(n)} + \mathbf{b}= \mathbf{u}_s^{(n)} - \mathbf{r}^{(n)}.
		\label{eq:Born_update}
\end{equation}
A Born iteration corresponds to a simple residual correction step. This generates the Neumann series expansion $\mathbf{u}_s = \sum_{i=1}^{\infty} \mathbf{A}^i \mathbf{u}_0$, with the convergence criterion of spectral radius  $\rho(\mathbf{A}) < 1$. This condition guarantees that successive iterations of the operator decay in magnitude. In practice, this criterion is often violated in strongly scattering or high-contrast media, where multiple scattering effects dominate and higher-order terms grow, leading to divergence.

\subsection{Direct Grid Optimization}

To enable convergence regardless of the spectral radius $\rho(\mathbf{A})$, the problem can be framed as the minimization of the $\ell_2$ norm of the residual with gradient descent. The discrete GI loss function (scaled by $1/2$) is:
\begin{equation}
	\mathcal{L} = \frac{1}{2} \big\| (\mathbf{I} - \mathbf{A})\mathbf{u}_s - \mathbf{b} \big\|_2^2.
	\label{eq:l2_loss}
\end{equation}
Consider the simplest case where the ``neural network'' has no hidden layers, and the trainable parameters $\boldsymbol{\theta}$ directly represent the scattered wavefield values at the grid points, such that $\mathbf{u}_s = \boldsymbol{\theta}$. Applying gradient descent to \eqref{eq:l2_loss} with a learning rate $\eta$, the parameter update rule is driven by the complex-valued gradient
\begin{equation}
	\nabla_{\mathbf{u}_s} \mathcal{L} = (\mathbf{I} - \mathbf{A})^H \big( (\mathbf{I} - \mathbf{A})\mathbf{u}_s - \mathbf{b} \big) = (\mathbf{I} - \mathbf{A})^H \mathbf{r},
\end{equation}
where $(\cdot)^H$ denotes the conjugate transpose (Hermitian adjoint) operator. The resulting gradient descent update for the scattered wavefield is:
\begin{equation}
	\mathbf{u}_s^{(n+1)} = \mathbf{u}_s^{(n)} - \eta (\mathbf{I} - \mathbf{A})^H \mathbf{r}^{(n)},
	\label{eq:landweber_update}
\end{equation}
which is known as Landweber iteration. This scheme solves the normal equation
\begin{equation}
	(\mathbf{I} - \mathbf{A})^H(\mathbf{I} - \mathbf{A})\mathbf{u}_s = (\mathbf{I} - \mathbf{A})^H\mathbf{b}.
\end{equation}
The operator $(\mathbf{I} - \mathbf{A})^H(\mathbf{I} - \mathbf{A})$ is Hermitian positive semi-definite, and the iteration \eqref{eq:landweber_update} converges to the least-squares solution provided the learning rate satisfies $\eta < 2 / \sigma_{\max}^2$, where $\sigma_{\max}$ is the largest singular value of $(\mathbf{I} - \mathbf{A})$. This formulation enables convergence in highly scattering media. However, in weakly scattering media, its convergence rate is slower than that of \eqref{eq:Born_update} because the condition number of the system is squared. To overcome the slow convergence induced by this squared condition number, a preconditioning mechanism is required.

\subsection{Neural Network Optimization}

Let the scattered wavefield be parameterized by a non-linear deep neural network $\mathcal{N}_{\boldsymbol{\theta}}$, such that $\mathbf{u}_s = \mathcal{N}_{\boldsymbol{\theta}}(\mathbf{y})$. The loss function remains unchanged, but the gradient is now taken with respect to the network weights $\boldsymbol{\theta}$. Applying the chain rule, we obtain
\begin{equation}
	\nabla_{\boldsymbol{\theta}} \mathcal{L} = \mathbf{J}_{\boldsymbol{\theta}}^H (\mathbf{I} - \mathbf{A})^H \mathbf{r},
\end{equation}
where $\mathbf{J}_{\boldsymbol{\theta}} = \partial \mathcal{N}_{\boldsymbol{\theta}} / \partial \boldsymbol{\theta}$ is the Jacobian matrix of the network output with respect to its parameters. The gradient descent update for the network's weights is:
\begin{equation}
	\boldsymbol{\theta}^{(n+1)} = \boldsymbol{\theta}^{(n)} - \eta \mathbf{J}_{\boldsymbol{\theta}}^H (\mathbf{I} - \mathbf{A})^H \mathbf{r}^{(n)}.
	\label{eq:weight_update}
\end{equation}

To understand how this weight update affects the physical field, we apply a first-order Taylor expansion to the network output, $\Delta \mathbf{u}_s \approx \mathbf{J}_{\boldsymbol{\theta}} \Delta \boldsymbol{\theta}$. Substituting the weight update \eqref{eq:weight_update} into this expansion yields the effective update step for the scattered wavefield
\begin{equation}
	\mathbf{u}_s^{(n+1)} \approx \mathbf{u}_s^{(n)} - \eta \big( \mathbf{J}_{\boldsymbol{\theta}} \mathbf{J}_{\boldsymbol{\theta}}^H \big) (\mathbf{I} - \mathbf{A})^H \mathbf{r}^{(n)}.
	\label{eq:preconditioned_update}
\end{equation}

Comparing \eqref{eq:preconditioned_update} to \eqref{eq:landweber_update}, the neural network introduces the matrix $\mathbf{P}_{\boldsymbol{\theta}} = \mathbf{J}_{\boldsymbol{\theta}} \mathbf{J}_{\boldsymbol{\theta}}^H$. This matrix, related to the Neural Tangent Kernel (NTK) \citep{jacot2018neural}, functions as a dynamic preconditioner. Rather than updating grid points independently, $\mathbf{P}_{\boldsymbol{\theta}}$ couples the spatial updates according to the geometric priors of the network architecture. 

 Standard Multi-Layer Perceptrons (MLPs) exhibit a bias toward low-frequency functions, resulting in vanishing eigenvalues in $\mathbf{P}_{\boldsymbol{\theta}}$ for high-wavenumber components. Our employed architecture with sinusoidal encoder and sine activations shapes $\mathbf{P}_{\boldsymbol{\theta}}$. Because the derivatives of the sine activation are themselves periodic, the Jacobian $\mathbf{J}_{\boldsymbol{\theta}}$ becomes populated by oscillatory functions. Consequently, the spectrum of the NTK approximation $\mathbf{J}_{\boldsymbol{\theta}} \mathbf{J}_{\boldsymbol{\theta}}^H$ is shifted. Aligning the encoding frequency with the maximum wavenumber of the Helmholtz operator improves the representation power and effective conditioning for high-wavenumber components. This transforms the update rule \eqref{eq:preconditioned_update} into a targeted high-frequency preconditioner.
 
\section{Numerical Results}
\label{sec:results}

This section presents numerical experiments evaluating the Green--Integral (GI) training strategy and its hybrid GI+PDE extension for scattered wavefield reconstruction. Their performance is compared against conventional physics-informed neural networks (PINNs) using three benchmark velocity models: Marmousi, Overthrust, and Otway.

These benchmark models are widely used in seismic wave propagation studies and provide realistic two-dimensional spatial distributions of acoustic wave speed. They are selected because each presents distinct physical and numerical challenges for neural wave solvers: \textbf{Marmousi} contains complex geological structures and sharp velocity contrasts, producing complicated wave propagation with strong multipathing; \textbf{Overthrust} represents a large-scale thrust-fault setting with smoothed velocity contrasts; and \textbf{Otway} is dominated by dense subwavelength horizontal layering, which challenges the network to capture the cumulative phase and amplitude effects.
%

\textbf{Experimental Setup}:  We utilize a fully connected deep neural network comprising five hidden layers with sine activation functions. To enhance high-frequency representation, a sinusoidal positional encoding layer is applied to the input coordinates \citep{abedi2025gabor}. Since coordinates are normalized to the background wavelength, the background propagating wavenumber is $2\pi$; thus, the frequency band $K=3$ is selected to ensure the maximum encoded frequency  exceeds it.

Training is conducted using the Adam optimizer with an initial learning rate of $10^{-3}$, which decays exponentially to $3.4 \times 10^{-4}$ over the course of the optimization. For the Marmousi test case, we utilize 128 neurons per layer. Given the increased structural complexity and spatial extent of the Overthrust and Otway models, the layers' width is increased to 150 neurons for those experiments to provide sufficient representational capacity.

	\begin{figure*}[t]
	\newcommand{\figlabel}[1]{\footnotesize (#1)}
\centering
		\begin{subfigure}{0.32\textwidth}
				\stackinset{l}{-2pt}{t}{2pt}{\figlabel{a}}{\includegraphics[width=\textwidth]{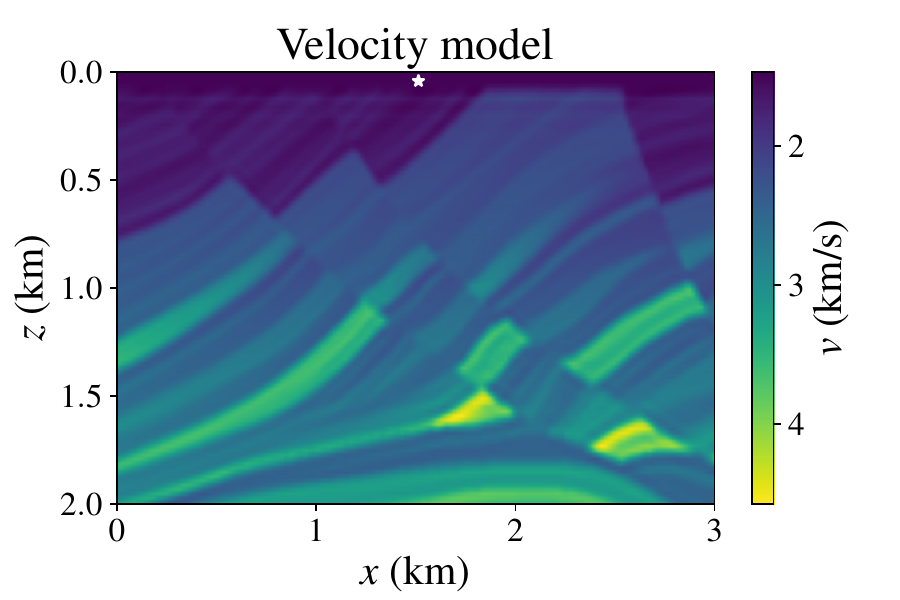}|
				}

		\end{subfigure}
		\begin{subfigure}{0.32\textwidth}
				\stackinset{l}{-2pt}{t}{2pt}{\figlabel{b}}{\includegraphics[width=\textwidth]{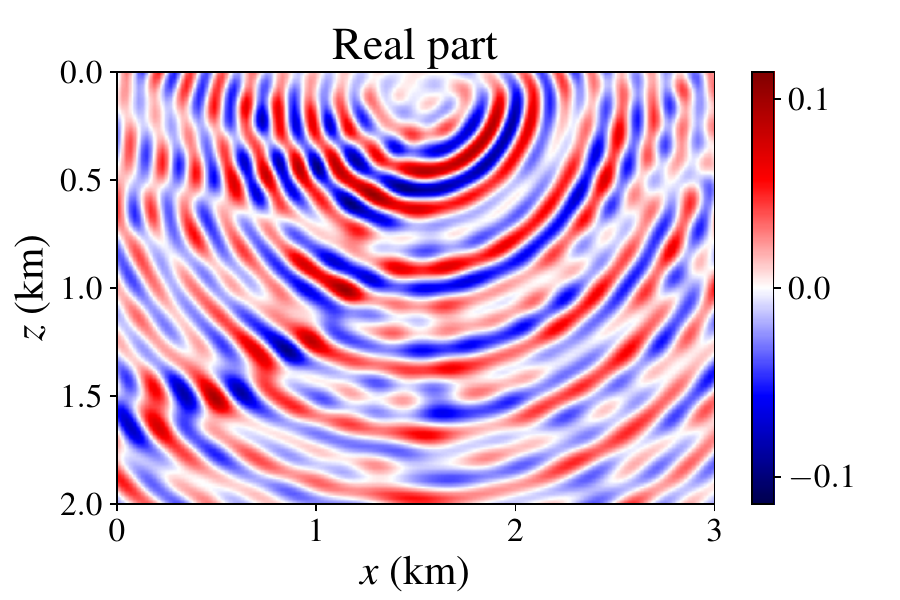}}

		\end{subfigure}
		\begin{subfigure}{0.32\textwidth}
				\stackinset{l}{-2pt}{t}{2pt}{\figlabel{c}}{\includegraphics[width=\textwidth]{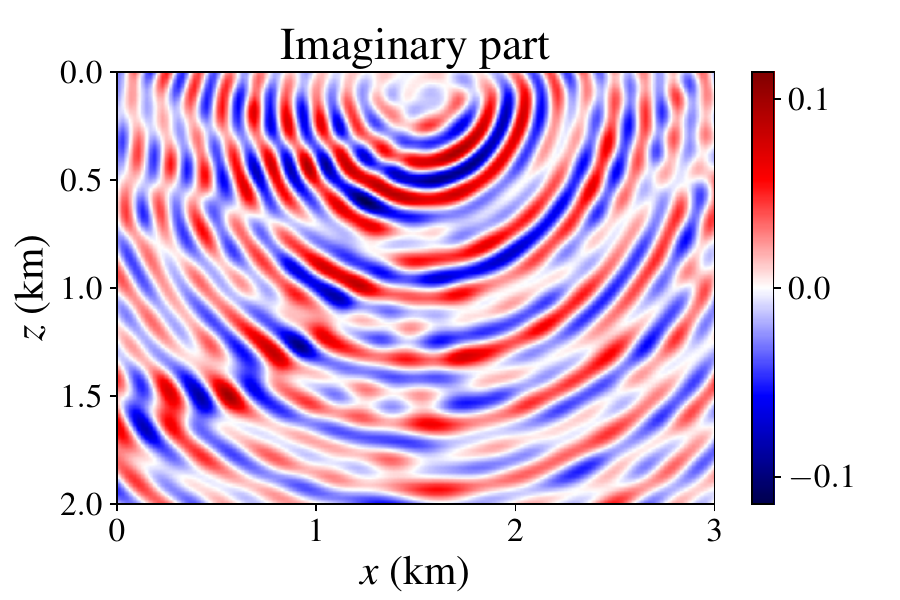}}

		\end{subfigure}
		
		\caption{Marmousi velocity model, and a 10 Hz finite-difference (FD) modeled scattered wavefield. This FD result is used as reference for validation of neural networks predictions.}
		\label{fig:marmousi_fd}
	\end{figure*}

	\subsection{Marmousi Velocity Model}
	The first experiment evaluates the GI neural network solver and its hybrid extension on a portion of the benchmark Marmousi velocity model  (Figure~\ref{fig:marmousi_fd}a). The Marmousi model is a realistic and widely used synthetic benchmark in seismic imaging, originally introduced by \cite{brougois1990Marmousi} to represent the complex geology of the Cuanza Basin. It is characterized by strong velocity variations, stratigraphic complexity, and sharp contrasts. Such heterogeneity makes it a challenging test case for wavefield simulation methods.
	
	We consider a selected region of the model and simulate a 10~Hz scattered wavefield using finite-difference (FD) modeling.  The FD solution is computed on a high-resolution grid and subsequently resampled to a $100 \times 150$ ($z \times x$) grid to serve as the reference for neural network validation (Figures~\ref{fig:marmousi_fd}).  
	
	Table~\ref{tab:gpu_memory_training_time1} summarizes the training metrics. It shows the number of integration points used in the GI loss ($N_y$), the number of collocation points used in the PDE loss ($N_x$), the peak GPU memory usage, the total training time on an NVIDIA L40S GPU, and the normalized mean squared error (NMSE) of the final prediction relative to the FD reference. Peak GPU memory usage, as reported by TensorFlow's internal allocator, includes allocations for model parameters, activations, gradients, and optimizer states, and thus provides a consistent measure of relative computational demand across experiments.

	\paragraph{Green--Integral  Training}
	
\begin{figure*}[t]
		\centering
		\begin{subfigure}{0.32\textwidth}
			\includegraphics[width=\textwidth]{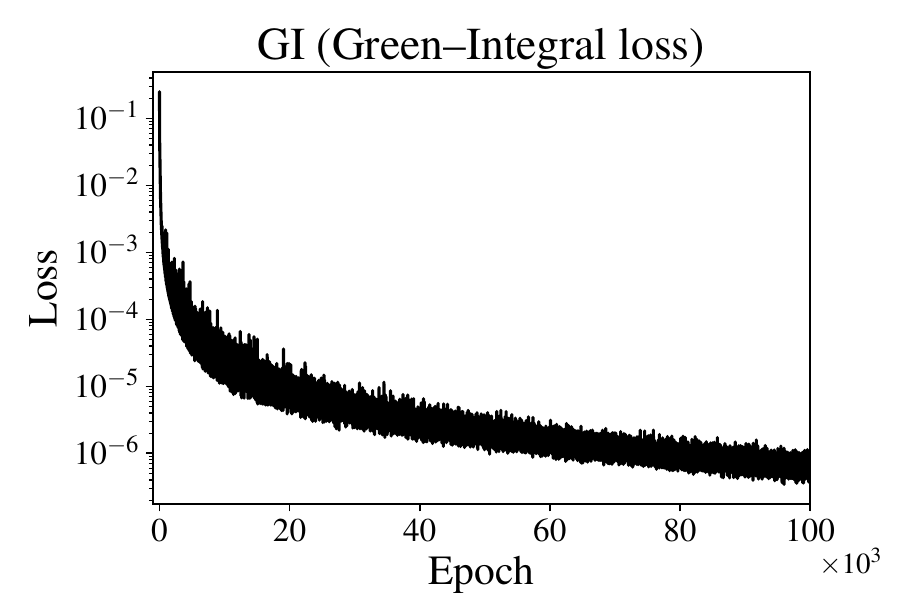}\\
			\includegraphics[width=\textwidth]{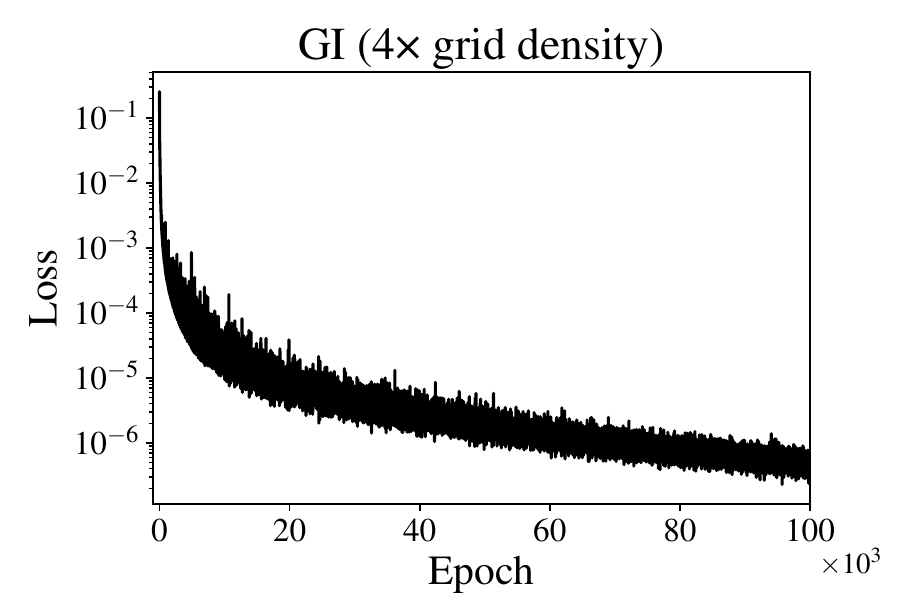}\\
			\includegraphics[width=\textwidth]{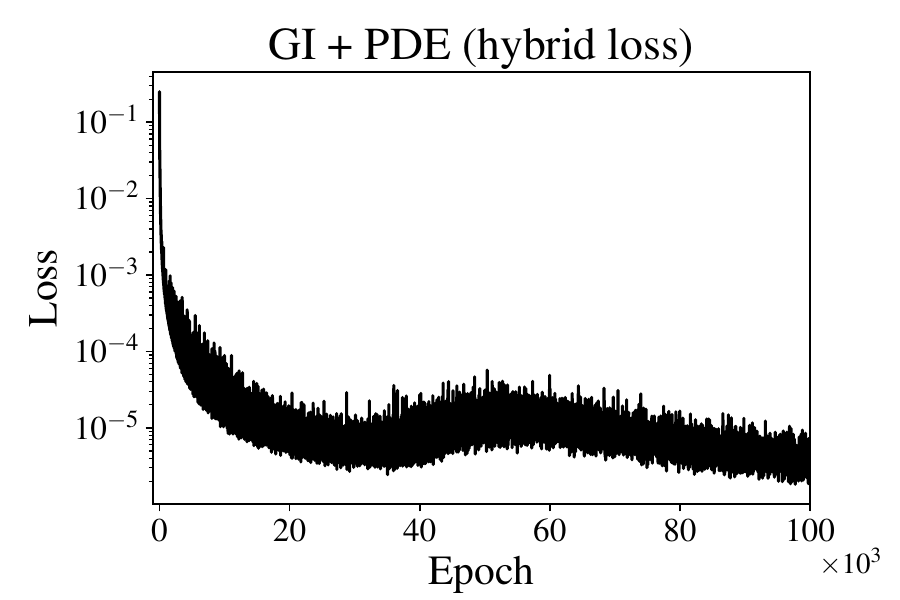}
			\caption{Loss}
		\end{subfigure}
		\begin{subfigure}{0.32\textwidth}
			\includegraphics[width=\textwidth]{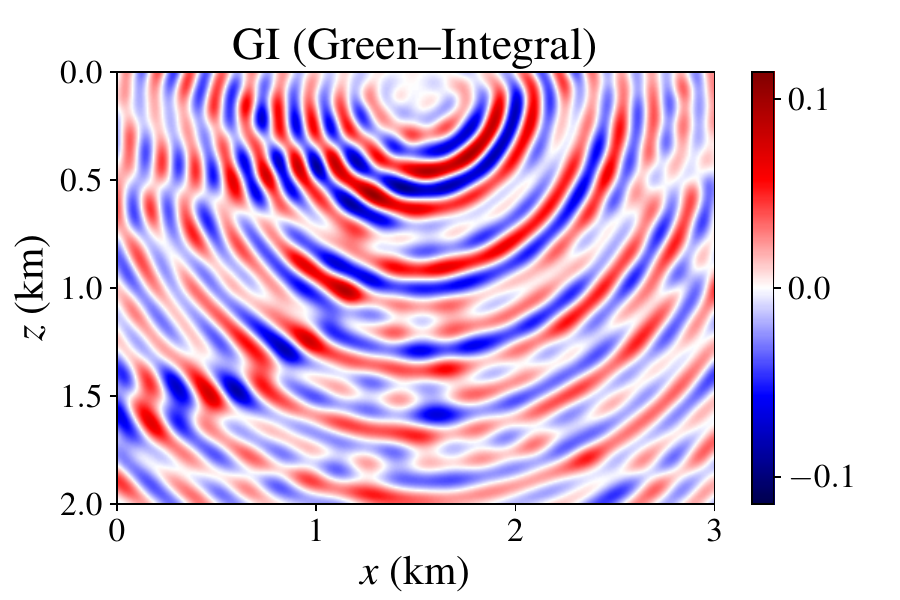}\\
			\includegraphics[width=\textwidth]{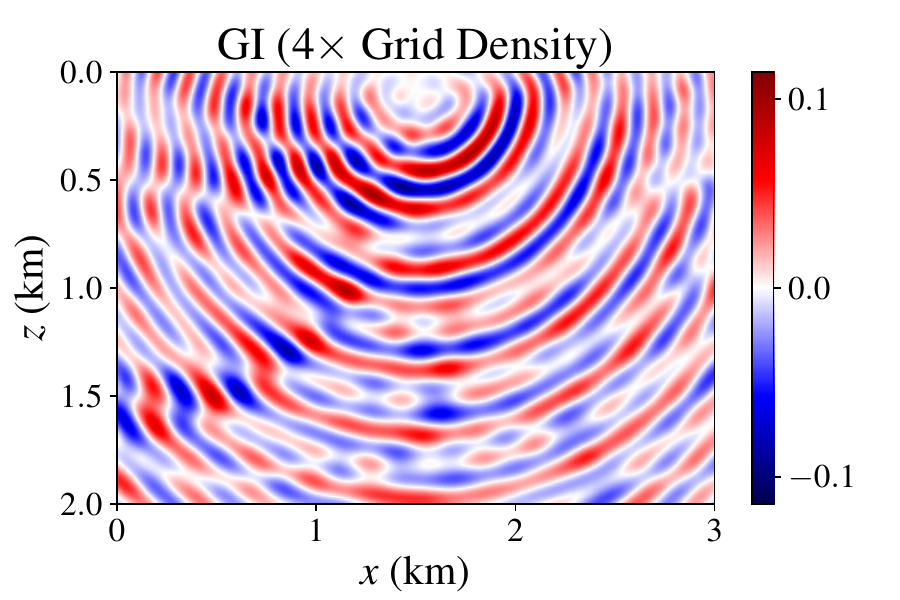}\\
			\includegraphics[width=\textwidth]{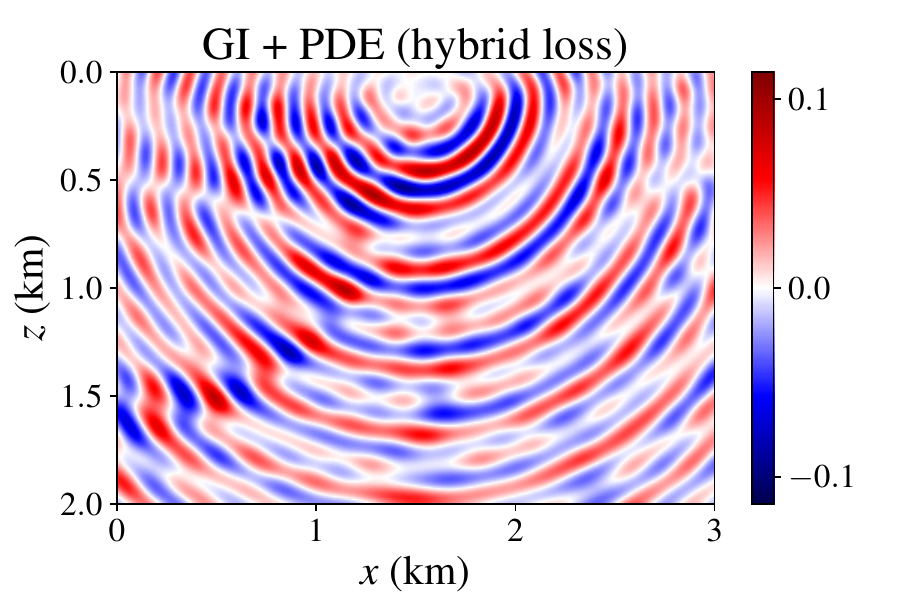}
			\caption{Prediction}
		\end{subfigure}
		\begin{subfigure}{0.32\textwidth}
			\includegraphics[width=\textwidth]{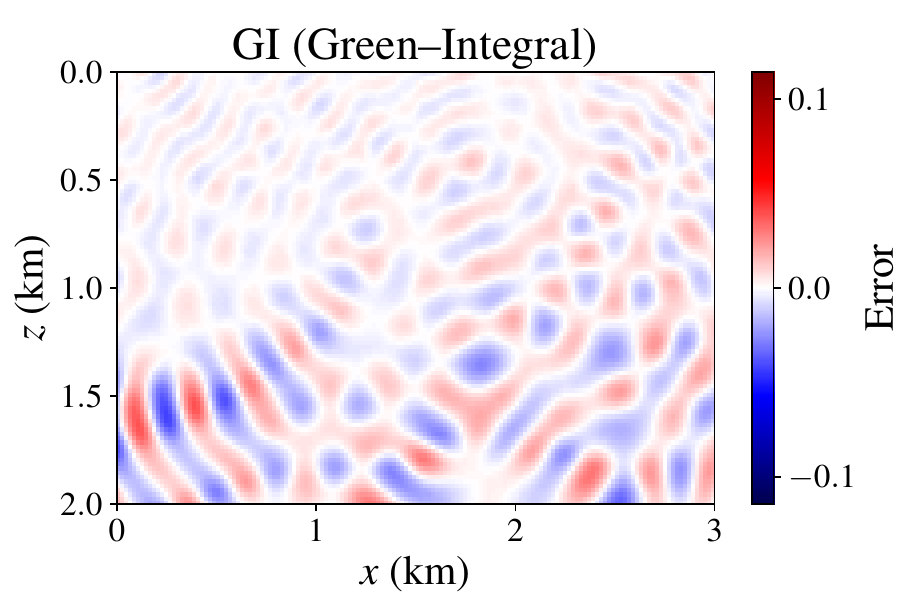}\\
			\includegraphics[width=\textwidth]{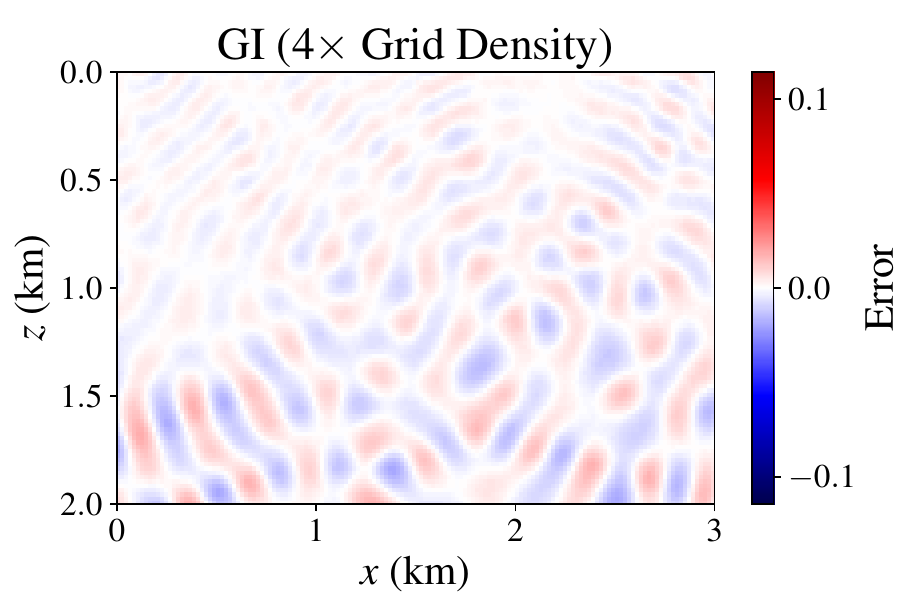}\\
			\includegraphics[width=\textwidth]{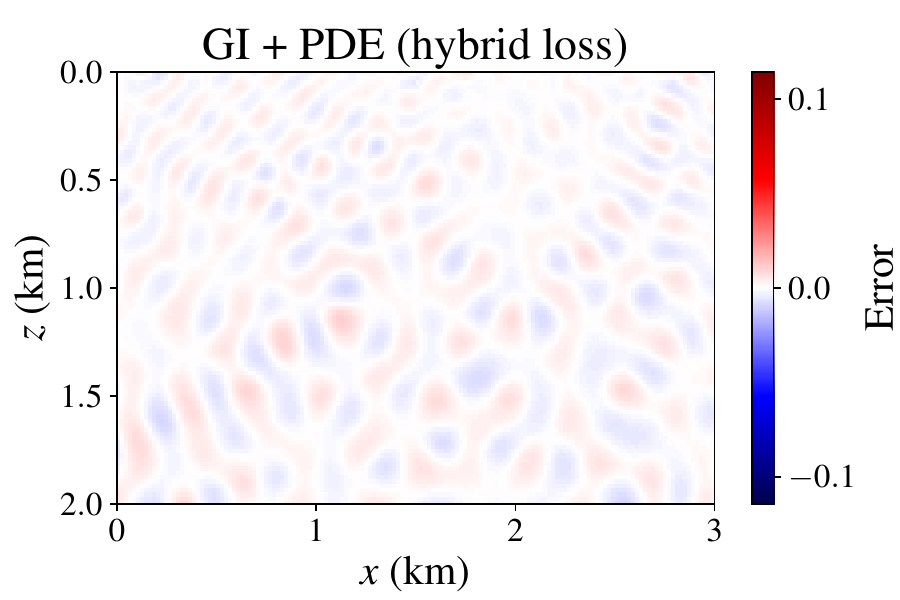}
			\caption{Error}
		\end{subfigure}
		
\caption{
	Training dynamics and reconstruction accuracy for the Marmousi model using Green--Integral (GI)-based formulations.
Columns show the loss evolution, the real part of the predicted scattered wavefield, and the absolute error with respect to the finite-difference reference solution (the imaginary parts have similar accuracy). Refinning the GI grid reduces the error; but the hybrid GI+PDE formulation shows the lowest errors.
}
		\label{fig:marmousi_proposed}
	\end{figure*}

	We begin by testing our GI neural network solver. Following the steps in the Implementation section, we train the network using the GI loss term (equation \eqref{eq:GI_loss}), evaluated on a fixed uniform grid of \( 120 \times 170 \) points with \(N_y = 20{,}400\). This GI formulation enforces global consistency between the predicted scattered field and its Green--integral reconstruction, coupling the solution at every spatial location to the entire domain. Figure~\ref{fig:marmousi_proposed} illustrates the evolution of the loss during training, the real part of the predicted scattered wavefield after 100{,}000 epochs, and the corresponding error maps for the different GI-based configurations.   
	
	To assess the impact of GI discretization on reconstruction accuracy, we repeat the experiment using a refined integration grid of \( 240 \times 340 \) points. This increased grid density improves the numerical accuracy of the GI loss evaluation and leads to a noticeable reduction in reconstruction error, particularly in regions influenced by strong scattering. The corresponding results demonstrate that part of the residual error in the GI solution is attributable to the finite resolution of the integral discretization rather than limitations of the network representation.
	
	While the GI-trained model yields globally consistent solutions and accurate wavefield predictions, maximum errors are localized near strong scatterers in the Marmousi model.  To address these local inaccuracies, we use the hybrid GI+PDE  loss (equation \eqref{eq:total_loss}), in which the PDE residual is evaluated at a small, randomly varying set of collocation points with \(N_x = 2{,}000\). The sampling procedure is explained in the Implementation section. 
	The PDE term provides local physical corrections that are difficult to capture through the uniform integral constraint alone.
	
	During training, the GI term dominates the early stages, enforcing global structure and outgoing-wave behavior, while the PDE term is gradually introduced to refine local accuracy.  
	Specifically, the weighting coefficient of the PDE term is increased with the number of epochs using a sigmoid schedule with a maximum value of \(\lambda = 0.01\).  The results show that combining the global GI constraint with a small number of PDE-based local corrections yields the most accurate reconstruction, effectively reducing localized errors while preserving the global consistency enforced by the integral formulation.

		
		\begin{table}[]
			\centering
\caption{Performance comparison of NN solvers for the 10~Hz Marmousi scattered wavefield (see Figures~\ref{fig:marmousi_fd}--\ref{fig:mse_evolution}). $N_y$ and $N_x$ denote the number of regular grid points (GI loss) and non-uniform collocation points (PDE loss), respectively. Peak GPU memory and training times were recorded on a single NVIDIA L40S. Bold values indicate the best performance, including the lowest Normalized Mean Square Error (NMSE).}
			\setlength{\tabcolsep}{2pt}
				\begin{tabular}{lccccc}
					\hline
					\textbf{Method} & $\mathbf{N_x}$ & $\mathbf{N_y}$ & \textbf{GPU Mem.} & \textbf{Time} & \textbf{NMSE} \\
					\hline
					\vspace{-5pt}\\
					GI (Green--Integral loss)        & 0        & 20{,}400 &\textbf{0.13} GB&\textbf{ 8} min  & $0.079$ \\
					GI (4$\times$ grid density)      & 0        & 81{,}600 & 0.49 GB & 11 min  & $0.025$ \\
					GI + PDE (hybrid loss)           & 2{,}000  & 20{,}400 & 0.28 GB & 30 min  & $\textbf{0.008}$ \\
					PINN (unconstrained PDE)         & 2{,}000  & 0        & 0.19 GB & 27 min  & $0.279$ \\
					PINN (PDE + S. constraint)   & 20{,}400 & 0        & 1.90 GB & 48 min  & $0.192$ \\
					PINN (PDE + PML)                 & 40{,}000 & 0        & 3.70 GB & 103 min & $0.027$ \\
					\hline
				\end{tabular}
				\label{tab:gpu_memory_training_time1}
			\end{table}

\begin{figure*}[t]
	\centering
	\begin{subfigure}{0.32\textwidth}
		\includegraphics[width=\textwidth]{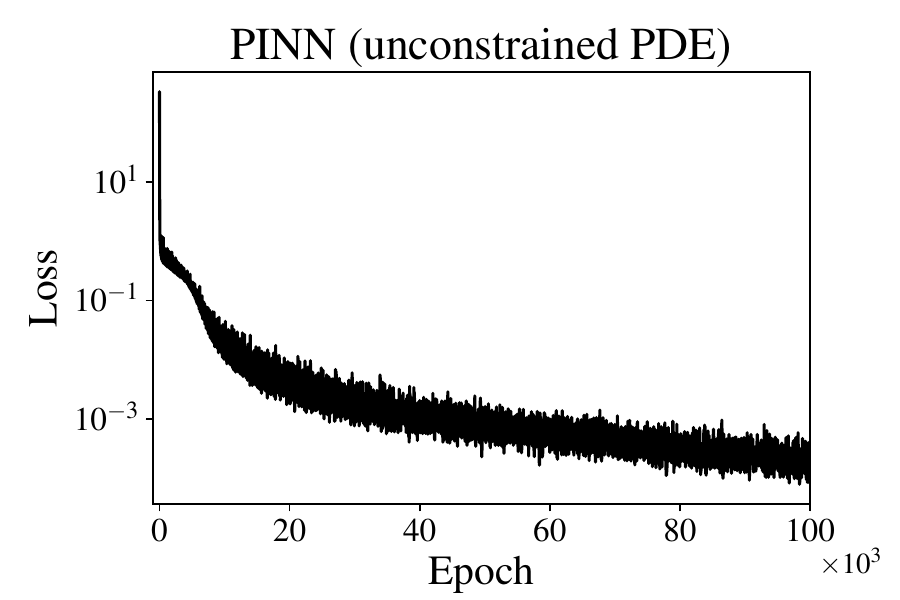}\\
		\includegraphics[width=\textwidth]{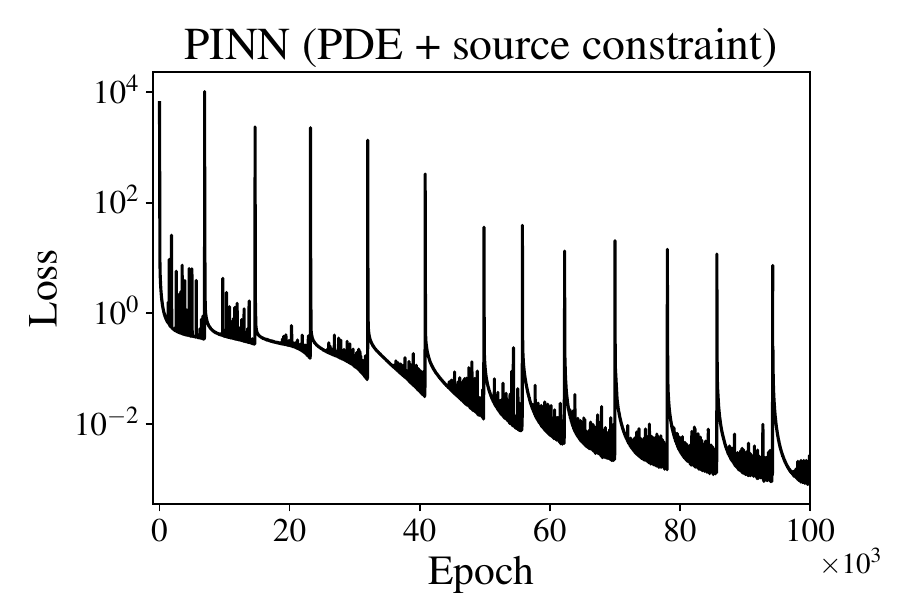}\\
		\includegraphics[width=\textwidth]{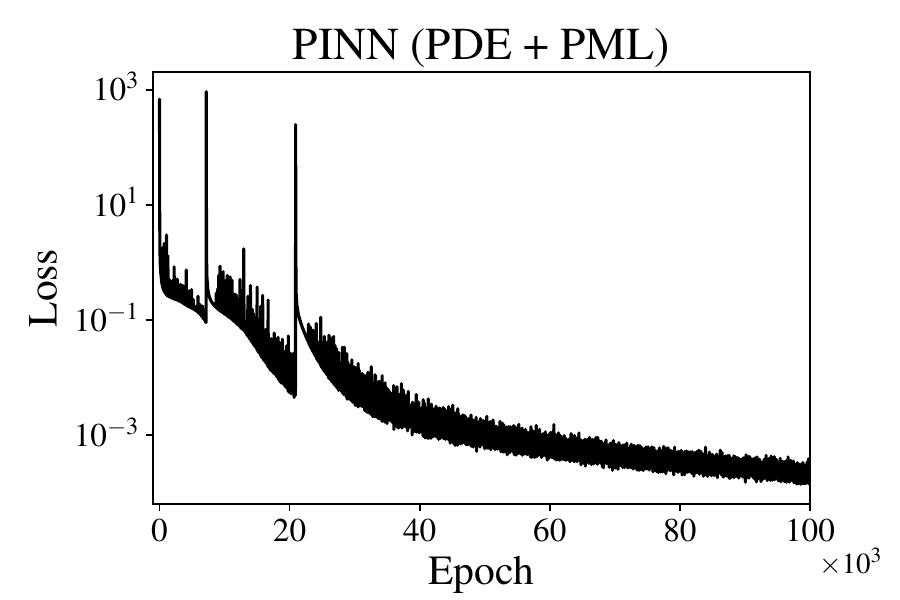}
		\caption{Loss}
	\end{subfigure}
	\begin{subfigure}{0.32\textwidth}
		\includegraphics[width=\textwidth]{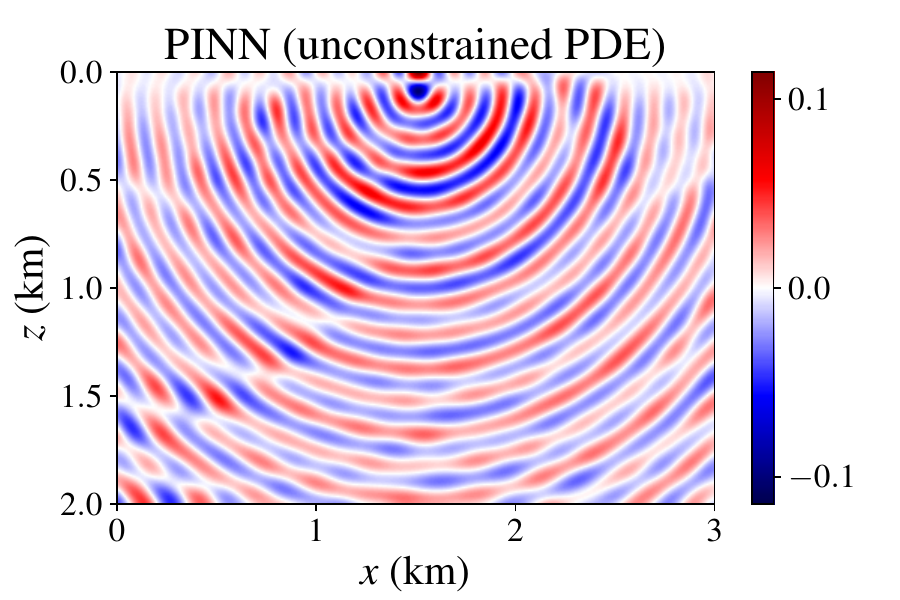}\\
		\includegraphics[width=\textwidth]{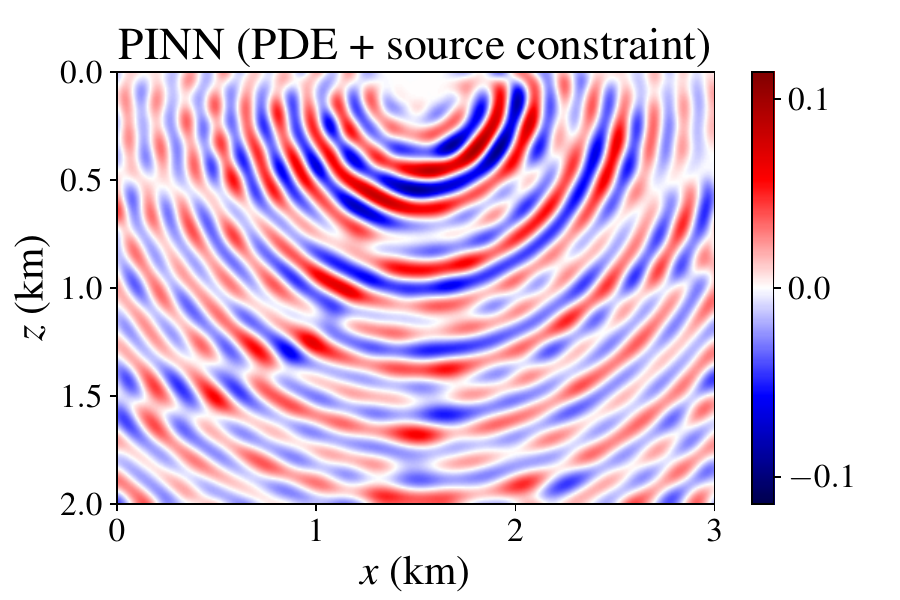}\\
		\includegraphics[width=\textwidth]{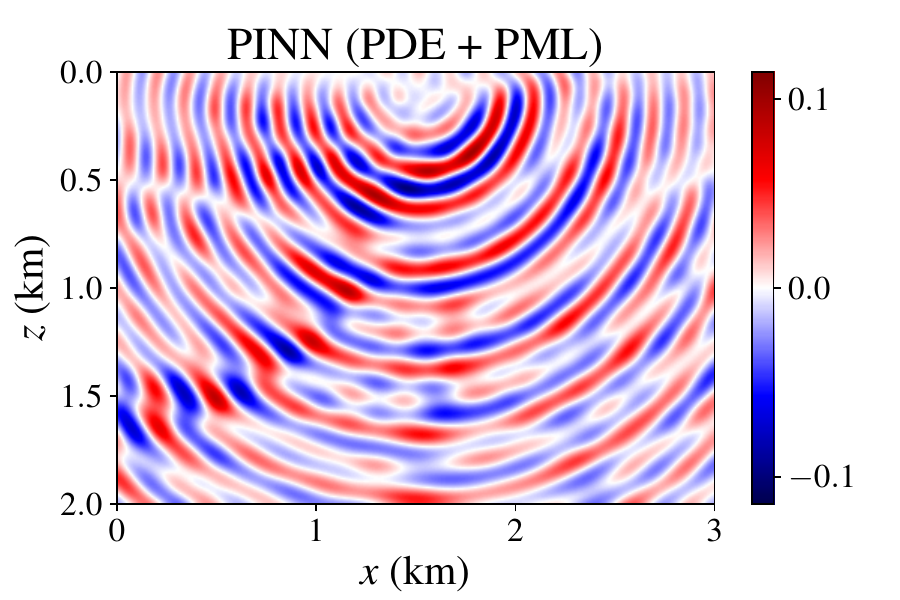}
		\caption{Prediction}
	\end{subfigure}
	\begin{subfigure}{0.32\textwidth}
		\includegraphics[width=\textwidth]{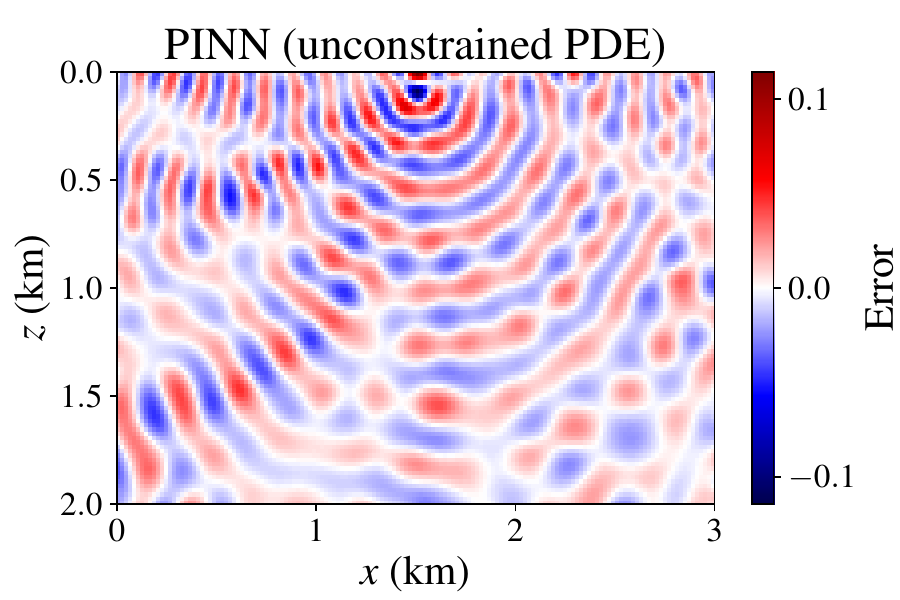}\\
		\includegraphics[width=\textwidth]{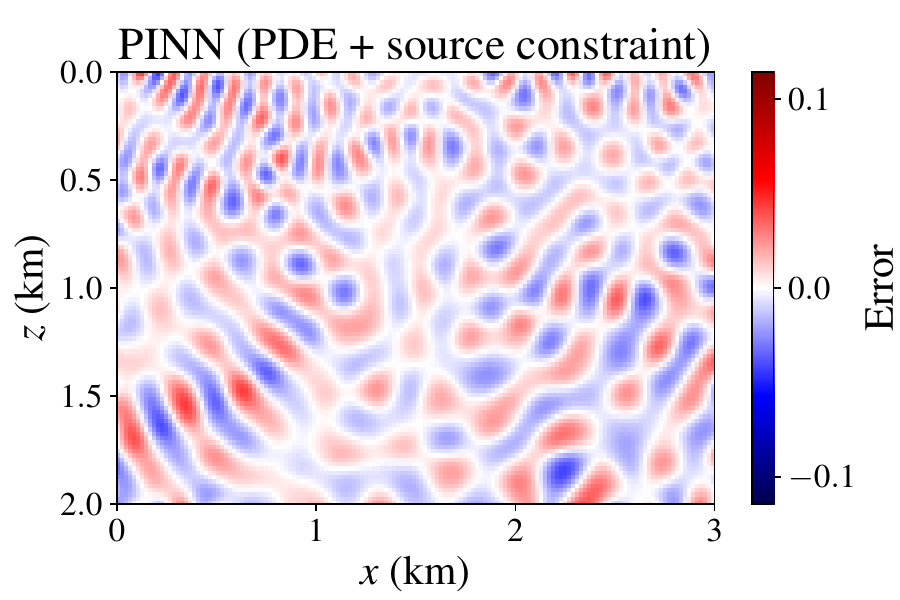}\\
		\includegraphics[width=\textwidth]{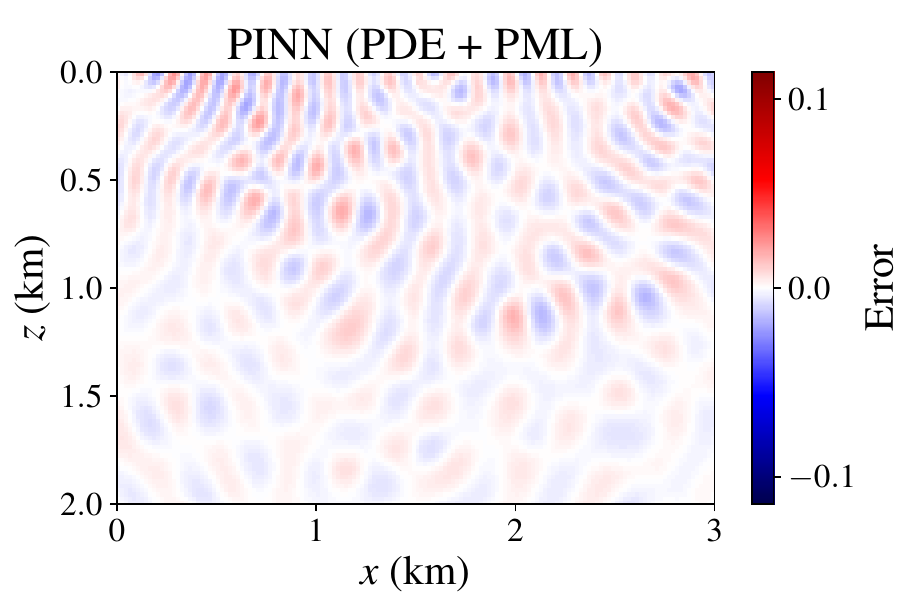}
		\caption{Error}
	\end{subfigure}
	
	\caption{
		Comparison of PINN-based formulations for the Marmousi model.
		Columns show the loss evolution, the real part of the predicted scattered wavefield, and the absolute error relative to the finite-difference reference. While additional constraints and absorbing layers improve stability and accuracy compared to the unconstrained formulation, they increase the computational costs and do not match the performance of the GI-based approaches shown in Fig.~\ref{fig:marmousi_proposed}.
	}
	\label{fig:training_comparison}
\end{figure*}

\begin{figure*}[]
	\newcommand{\figlabel}[1]{\footnotesize (#1)}
	\centering
	\begin{subfigure}{0.48\textwidth}
		\stackinset{l}{-2pt}{t}{2pt}{\figlabel{a}}{	\includegraphics[width=\textwidth]{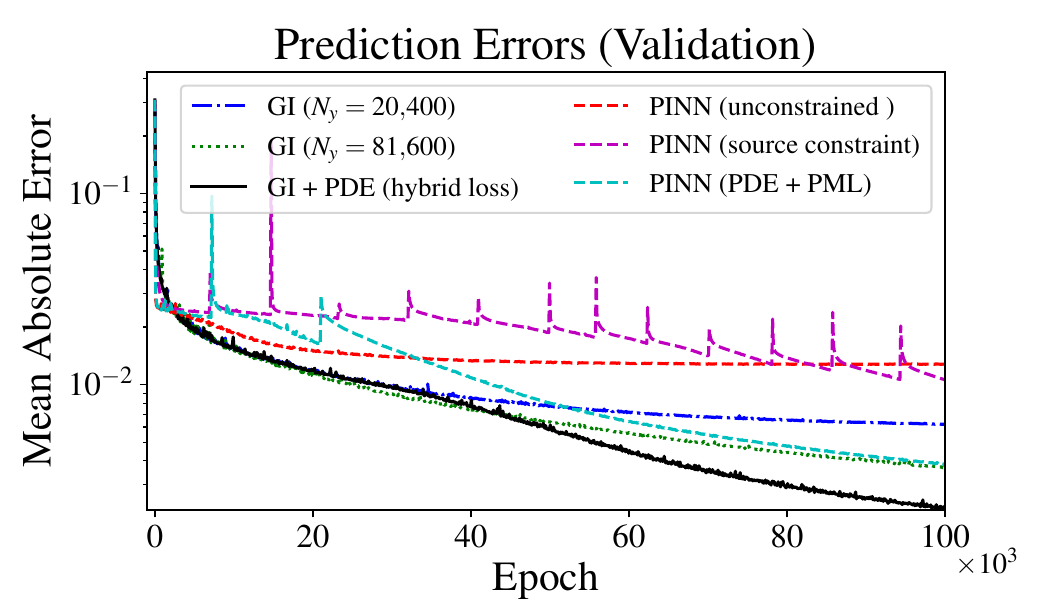}}
	\end{subfigure}
		\begin{subfigure}{0.48\textwidth}
			\stackinset{l}{-2pt}{t}{2pt}{\figlabel{b}}{\includegraphics[width=\textwidth]{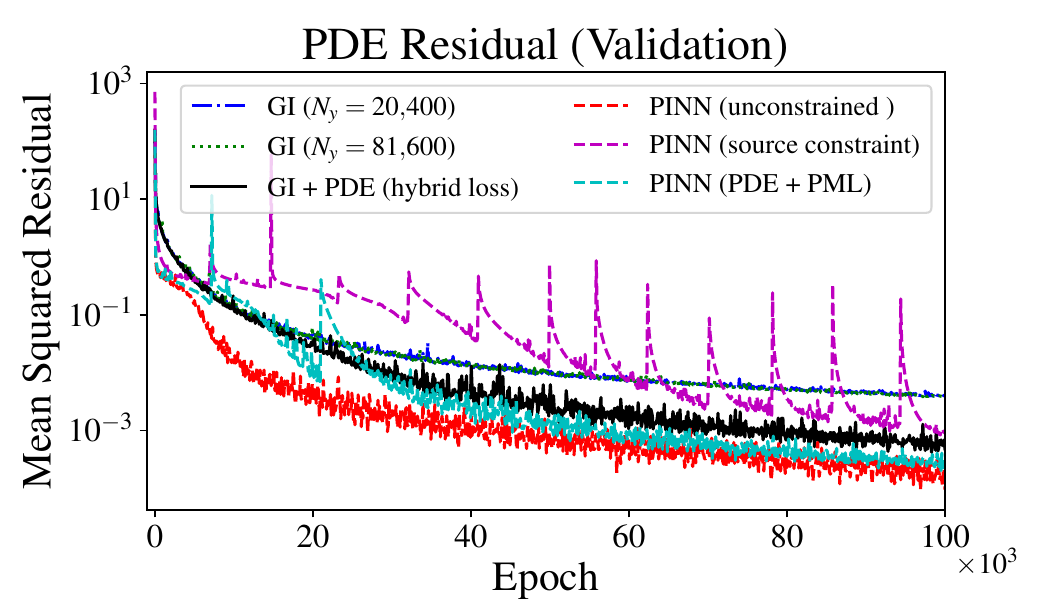}}
	\end{subfigure}
\caption{
 Evolution of (a) Mean Absolute Error (MAE) and (b) PDE residual during training, evaluated on a validation set of collocation points. The results correspond to the experiments illustrated in Figures~\ref{fig:marmousi_proposed} and~\ref{fig:training_comparison}.  }

	\label{fig:mse_evolution}
\end{figure*}

\paragraph{Comparison with PINN Formulations}

We next compare the GI-based approaches with several PINN formulations that rely on PDE-residual minimization for learning the scattered Helmholtz equation.
We begin with a baseline PINN trained only on the PDE residual, without additional boundary treatments or constraints. This experiment corresponds to a standard PINN formulation for the scattered Helmholtz equation, introduced by \cite{alkhalifah2021}. While such PDE-only formulations have been reported to perform adequately for simpler, smoothly varying velocity models, our results confirm that enforcing only local PDE consistency is insufficient to prevent convergence to nonphysical solutions in the presence of strong heterogeneities. The training fails, converging to a solution with large residuals and pronounced errors near the source region (Figure~\ref{fig:training_comparison}).

To softly constrain the admissible solution space, the source constraint introduced by \cite{Huang2023} penalizes the amplitude of the scattered field \(U_s\) within a region of approximately half a wavelength surrounding the source. This constraint is based on an assumption that the scattered field vanishes at the source location. However, in the Marmousi model considered here, this assumption is violated due to strong near-source reflections, which generate nonzero scattered energy in the vicinity of the source (Figure~\ref{fig:marmousi_fd}).
As shown in Figure~\ref{fig:training_comparison}, incorporating this constraint leads to unstable training behavior, as the network attempts to simultaneously satisfy the PDE residual and the artificial amplitude suppression near the source. For training, besides random collocation points that sample the domain uniformly, additional points are introduced within the constraint region to account for its limited spatial extent relative to the full domain. To facilitate comparison, the total number of points in this test matches that of the first GI test.

We also test a PINN formulation that incorporates a perfectly matched layer (PML) of thickness 0.5~km surrounding the computational domain, following the implementation of \cite{abedi2025gabor}.
The PML effectively absorbs outgoing waves and improves boundary behavior, resulting in more accurate wavefield reconstructions compared to the unconstrained and source-constrained PINN formulations (Figure~\ref{fig:training_comparison}).
However, this improvement comes at a high computational cost, as the enlarged domain and corresponding increased number of collocation points lead to higher memory usage and longer training times. Further comparison of the performance of all GI-based and PINN-based formulations is presented below.

\paragraph{Training Cost and Validation Accuracy}

A quantitative comparison of all six training configurations is summarized in Table~\ref{tab:gpu_memory_training_time1}, and their convergence behavior is illustrated in Figure~\ref{fig:mse_evolution}, which reports both the validation prediction error and the validation PDE residual as functions of training epoch.

The GI training requires the lowest computational cost, the minimum GPU memory, and the shortest training time. The refined-grid GI-trained model achieves a validation error comparable to that of the PINN with PML, while requiring approximately one-tenth of the training time and a fraction of the memory usage of the PINN. This result highlights the efficiency of the GI formulation for accurate wavefield reconstructions without explicit absorbing boundaries.

The PDE-residual-based PINN formulations converge more slowly.  
The unconstrained and the source-constrained variants stall at higher validation error levels, reflecting the difficulty of restricting the admissible solution space without proper boundary conditions. The PML-based PINN effectively enforces outgoing-wave behavior and improves reconstruction accuracy, but at a substantial computational cost: both GPU memory usage and training time are several times higher than those required by the GI-based approaches.

An observation from Figure~\ref{fig:mse_evolution} is that minimizing the PDE residual alone does not guarantee convergence to the correct physical solution.  
The PINN with unconstrained PDE attains the lowest validation PDE residual among all methods, yet produces the largest mismatch with the FD reference wavefield.  

Among all tested methods, the hybrid GI+PDE formulation achieves the lowest error while maintaining low GPU memory usage and moderate training time.  Its prediction error decreases rapidly during the early stages of training and shows stable and efficient convergence. The hybrid loss training shows an effective balance between global physical consistency and localized corrections required for this velocity model with sharp, compact scatterers.

\subsection{Overthrust Velocity Model}
\begin{figure*}[t]
	\centering
	\newcommand{\figlabel}[1]{\footnotesize (#1)}
	
	\begin{subfigure}{0.32\textwidth}
		\centering
		\stackinset{l}{-2pt}{t}{2pt}{\figlabel{a}}{%
			\includegraphics[width=\textwidth]{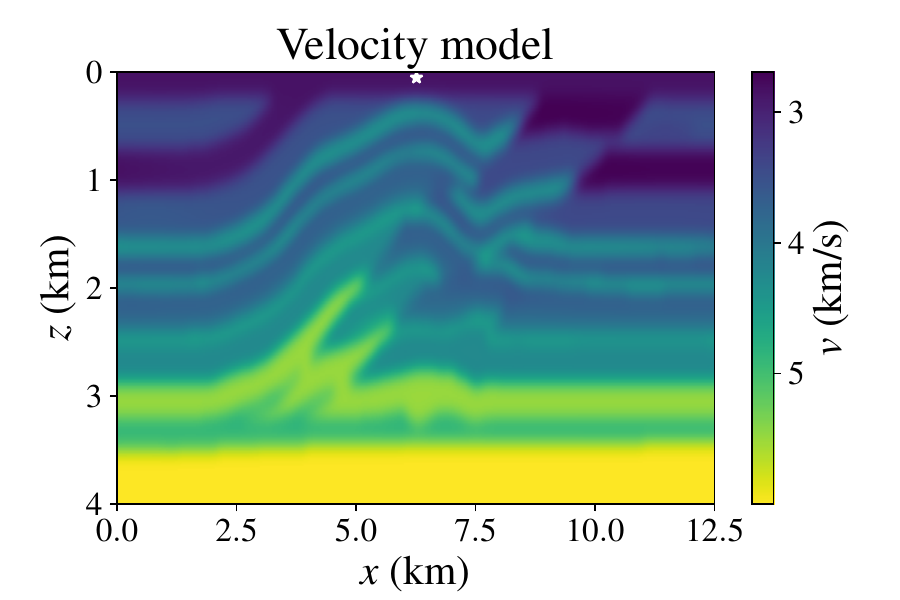}}
		\label{fig:OVE_velocity}
	\end{subfigure}
\begin{subfigure}{0.32\textwidth}
	\centering
	\stackinset{l}{-2pt}{t}{2pt}{\figlabel{b}}{%
		\includegraphics[width=\textwidth]{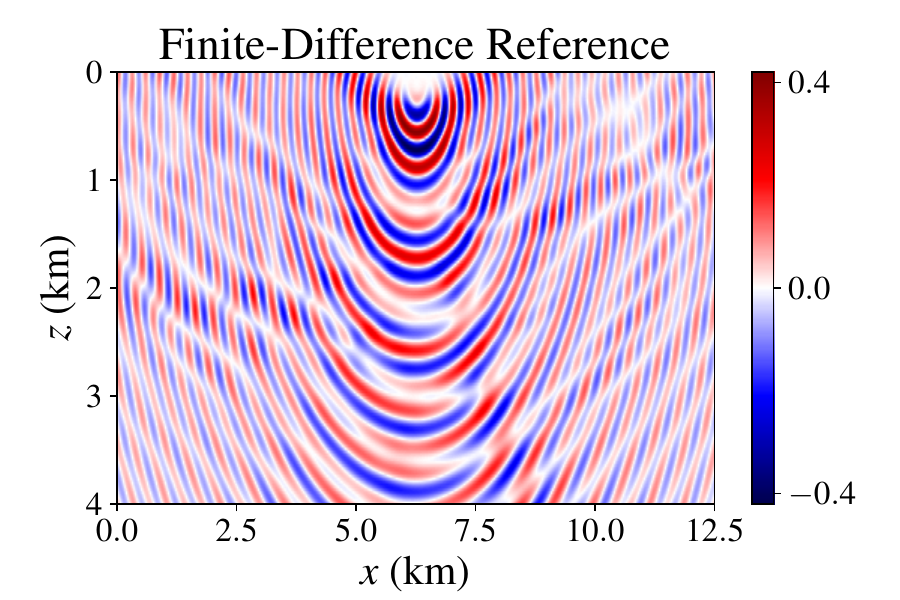}}
	\label{fig:OVE_fd}
\end{subfigure}
\begin{subfigure}{0.32\textwidth}
	\centering
	\stackinset{l}{-2pt}{t}{2pt}{\figlabel{c}}{%
		\includegraphics[width=\textwidth]{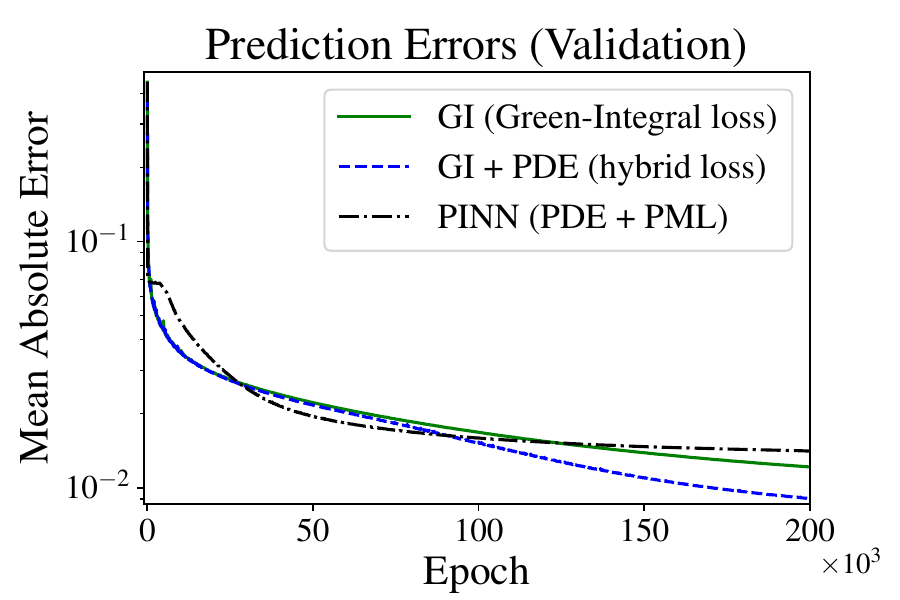}}
	\label{fig:OVE_val_error}
\end{subfigure}
\\
\begin{subfigure}{0.32\textwidth}
	\centering
	\stackinset{l}{-2pt}{t}{2pt}{\figlabel{d}}{%
		\includegraphics[width=\textwidth]{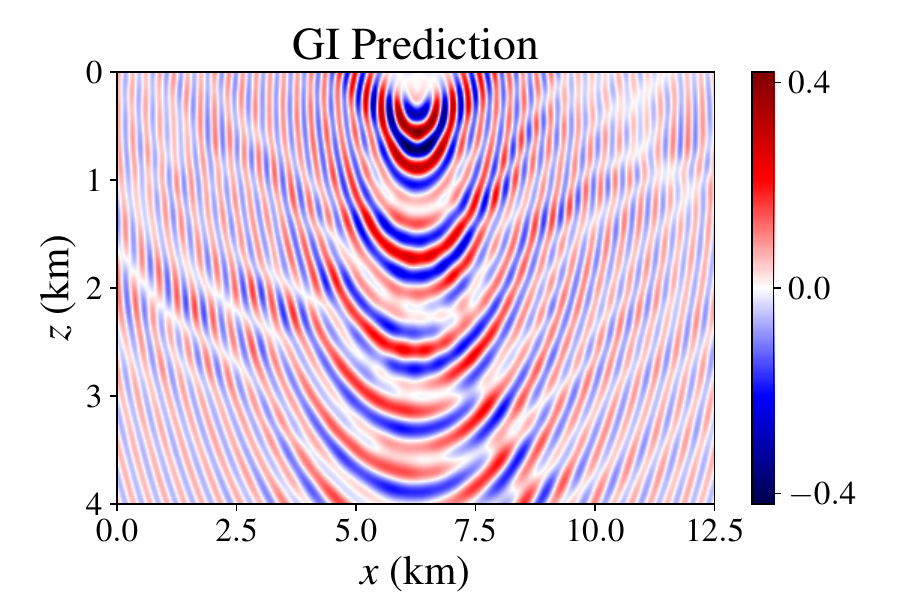}}
	\label{fig:OVE_prediction_GI}
\end{subfigure}
\begin{subfigure}{0.32\textwidth}
	\centering
	\stackinset{l}{-2pt}{t}{2pt}{\figlabel{e}}{%
		\includegraphics[width=\textwidth]{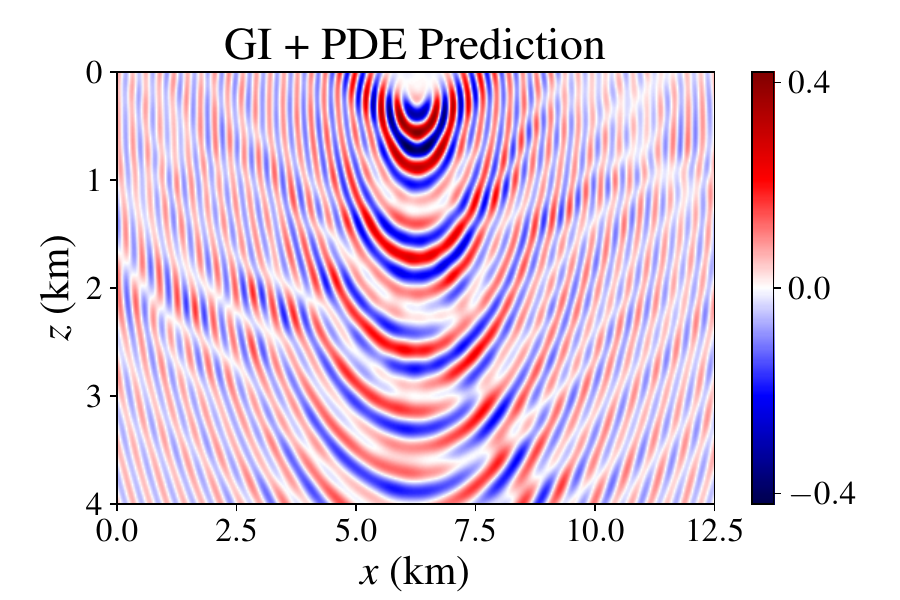}}
	\label{fig:OVE_prediction_GIPDE}
\end{subfigure}
\begin{subfigure}{0.32\textwidth}
	\centering
	\stackinset{l}{-2pt}{t}{2pt}{\figlabel{f}}{%
		\includegraphics[width=\textwidth]{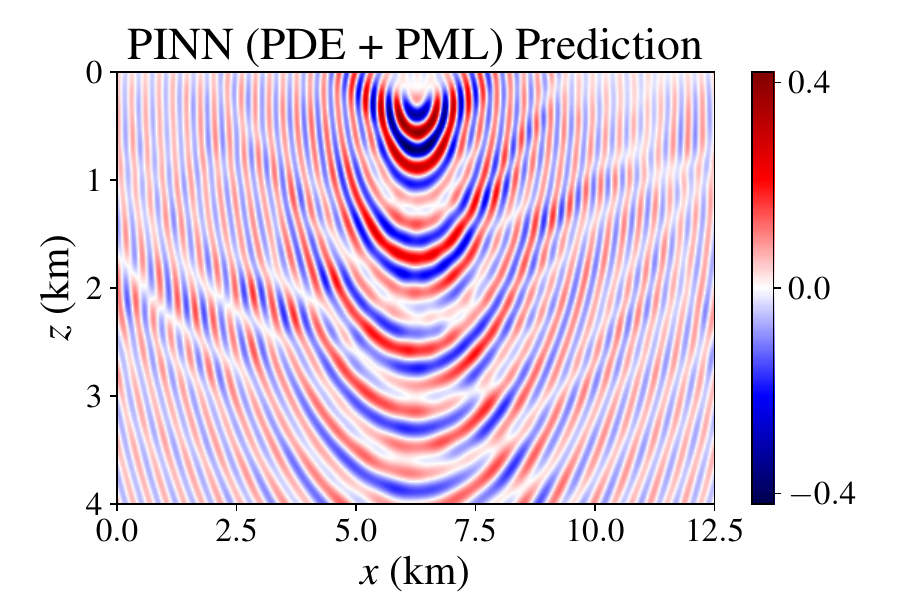}}
	\label{fig:OVE_prediction_PDEPML}
\end{subfigure}
\\
\begin{subfigure}{0.32\textwidth}
	\centering
	\stackinset{l}{-2pt}{t}{2pt}{\figlabel{g}}{%
		\includegraphics[width=\textwidth]{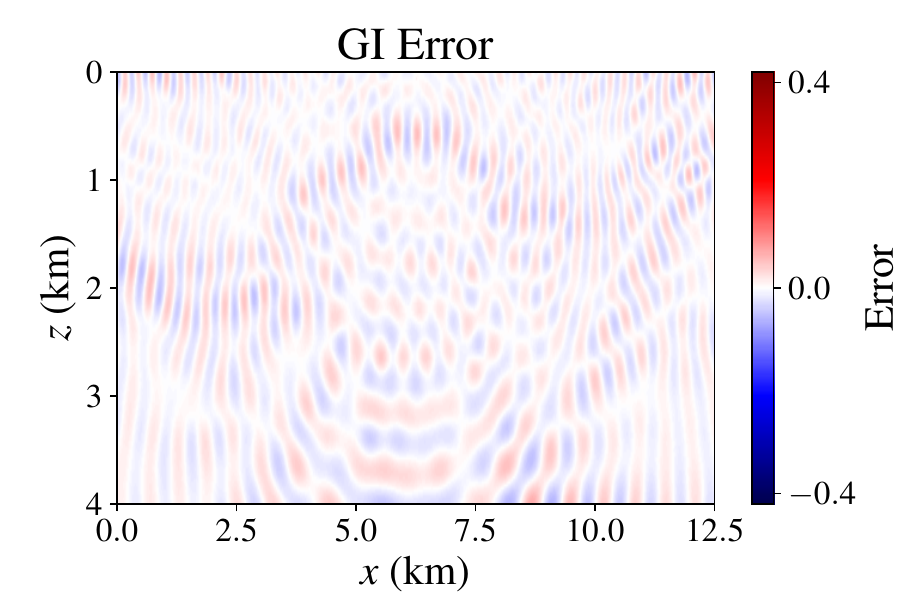}}
	\label{fig:OVE_prediction_GI_error}
\end{subfigure}
\begin{subfigure}{0.32\textwidth}
	\centering
	\stackinset{l}{-2pt}{t}{2pt}{\figlabel{h}}{%
		\includegraphics[width=\textwidth]{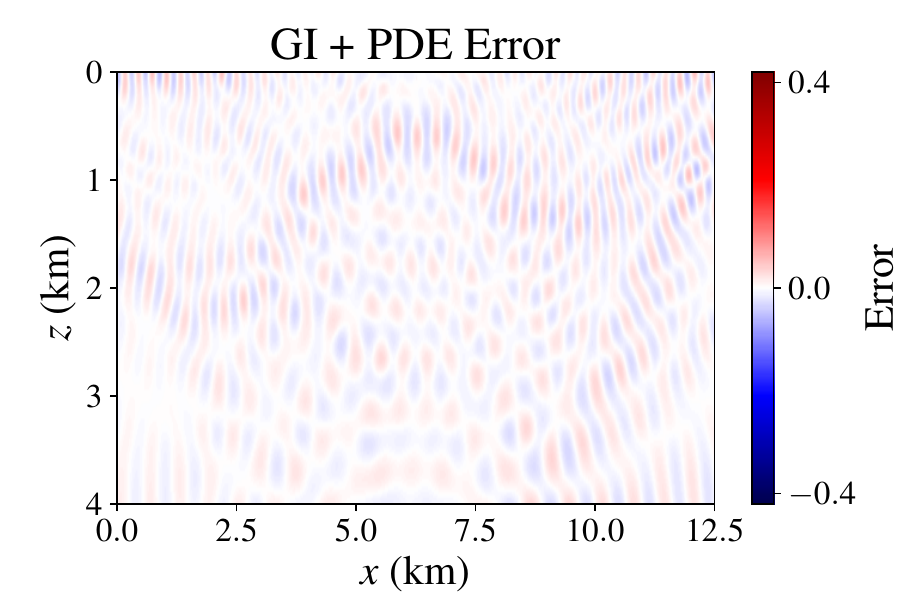}}
	\label{fig:OVE_prediction_GIPDE_error}
\end{subfigure}
\begin{subfigure}{0.32\textwidth}
	\centering
	\stackinset{l}{-2pt}{t}{2pt}{\figlabel{i}}{%
		\includegraphics[width=\textwidth]{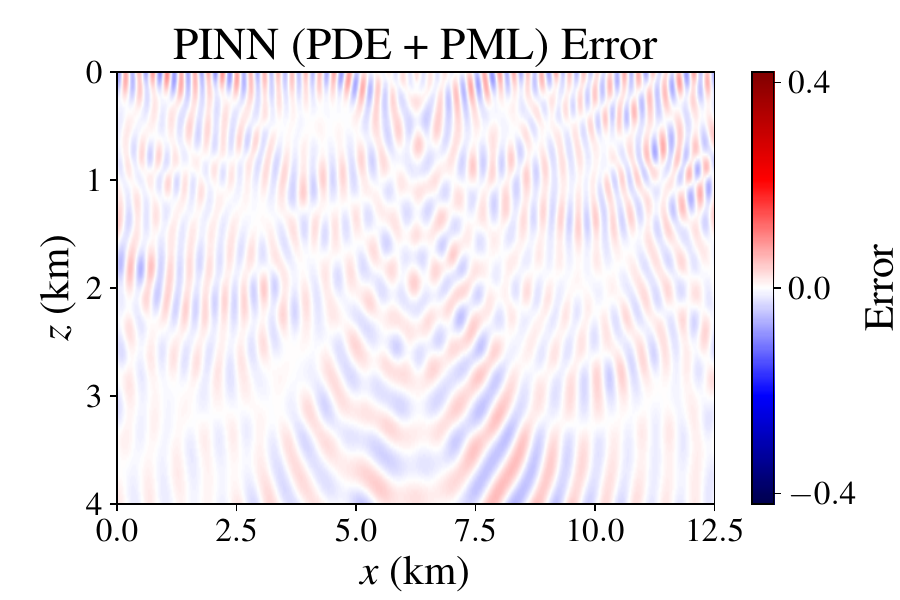}}
	\label{fig:OVE_prediction_PDEPML_error}
\end{subfigure}
\\
	\caption{
	Results for the larger Overthrust velocity model at 10~Hz.
	The GI-trained model achieves a similar accuracy to the PINN formulation at substantially lower computational cost (see Table \ref{tab:overthrust_cost}), while the hybrid GI+PDE loss yields the most accurate reconstruction.
}

	\label{fig:overthrust_results}
\end{figure*}

We next evaluate the performance of the GI--based training strategies on the smoothed Overthrust velocity model \citep{aminzadeh1994seg,alkhalifah2021}, a synthetic benchmark that shares structural similarities with the subsurface of the Canadian foothills. 
The model is characterized by two large overthrust faults and laterally extensive structures, resulting in a domain that is significantly larger than the Marmousi model but comparatively smoother, with fewer sharp small-scale heterogeneities (Figure~\ref{fig:overthrust_results}a).
A 10~Hz scattered wavefield is simulated using FD modeling and presented on a $160 \times 500$ grid to serve as the reference solution for validation. 
For this experiment, we use a larger neural network architecture with five hidden layers of 150 neurons. All remaining network and optimization settings are kept identical to those used in the previous experiments.

Figure~\ref{fig:overthrust_results} summarizes the results for this test.  
The first row shows the Overthrust velocity model, the FD reference scattered wavefield, and the evolution of the validation error during training.  
The second row compares the predicted scattered wavefields obtained using GI training, hybrid GI+PDE training, and a PINN formulation based on the PDE residual with a PML of 250~m thickness.  
The corresponding error maps are shown in the third row. While all three approaches achieve comparable predictions, their training costs differ substantially.

A quantitative comparison of training cost and accuracy is reported in Table~\ref{tab:overthrust_cost}.  Using the same number of training points, the GI-based model and the PINN with PML exhibit comparable validation accuracy throughout training, while we see a 20-fold reduction in training time and a 15-fold reduction in peak GPU memory usage in our GI training. Although the hybrid GI+PDE model incurs a moderate increase in computational cost relative to GI-only training, the added local correction further improves accuracy, confirming that PDE-based refinement remains beneficial in this smoother velocity model.

\begin{table}[]
	\centering
	\caption{Comparison of NN training methods for simulating the 10~Hz scattered wavefield in the Overthrust velocity model.}
	\label{tab:overthrust_cost}
	\setlength{\tabcolsep}{3pt} 
	\begin{tabular}{lccccc}
		\hline
		\textbf{Method} & $\mathbf{N_x}$ & $\mathbf{N_y}$ & \textbf{GPU Mem.} & \textbf{Time} & \textbf{NMSE} \\
		\hline
		\vspace{-8pt}\\
		GI (Green--Integral loss)    & 0        & 95{,}004 & \textbf{0.68} GB & \textbf{28} min   & $0.030$ \\
		GI + PDE (hybrid loss)    & 4{,}750  & 95{,}004 & 1.11 GB          & 75 min  & $\textbf{0.018}$ \\
		PINN (PDE + PML)        & 95{,}004 & 0        & 10.19 GB         & 560 min & $0.042$ \\
		\hline
	\end{tabular}
\end{table}
	
\subsection{Otway Velocity Model}

\begin{figure*}[t]
	\centering
	\newcommand{\figlabel}[1]{\footnotesize (#1)}
	
	\begin{subfigure}{0.32\textwidth}
		\centering
		\stackinset{l}{-2pt}{t}{2pt}{\figlabel{a}}{%
			\includegraphics[width=\textwidth]{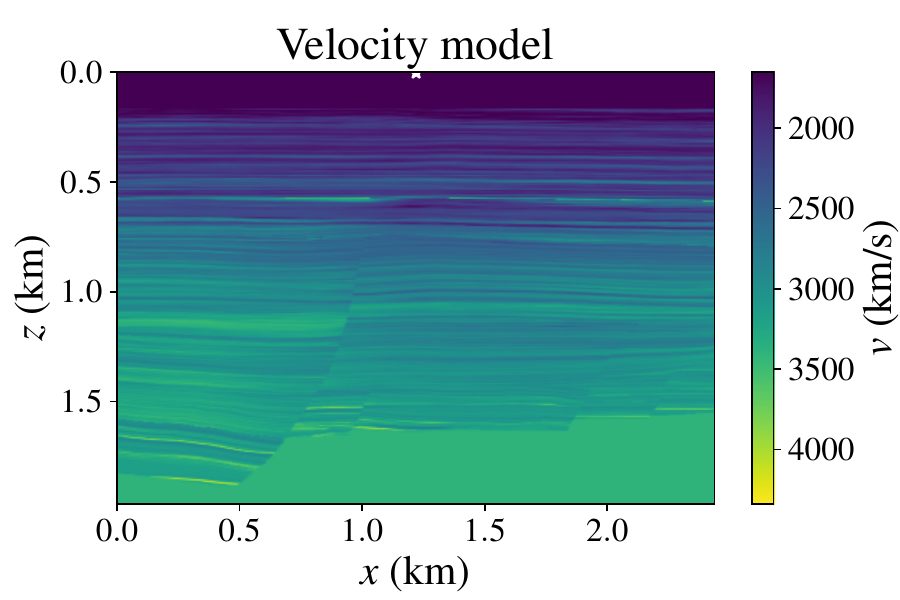}}
		\label{fig:otway_velocity}
	\end{subfigure}
	\begin{subfigure}{0.32\textwidth}
		\centering
		\stackinset{l}{-2pt}{t}{2pt}{\figlabel{b}}{%
			\includegraphics[width=\textwidth]{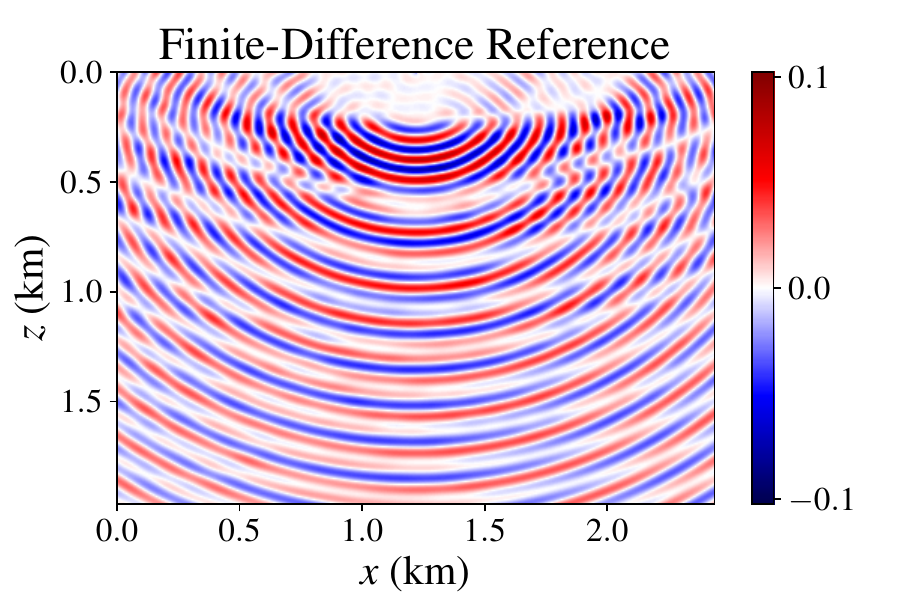}}
		\label{fig:otway_fd}
	\end{subfigure}
	\begin{subfigure}{0.32\textwidth}
		\centering
		\stackinset{l}{-2pt}{t}{2pt}{\figlabel{c}}{%
			\includegraphics[width=\textwidth]{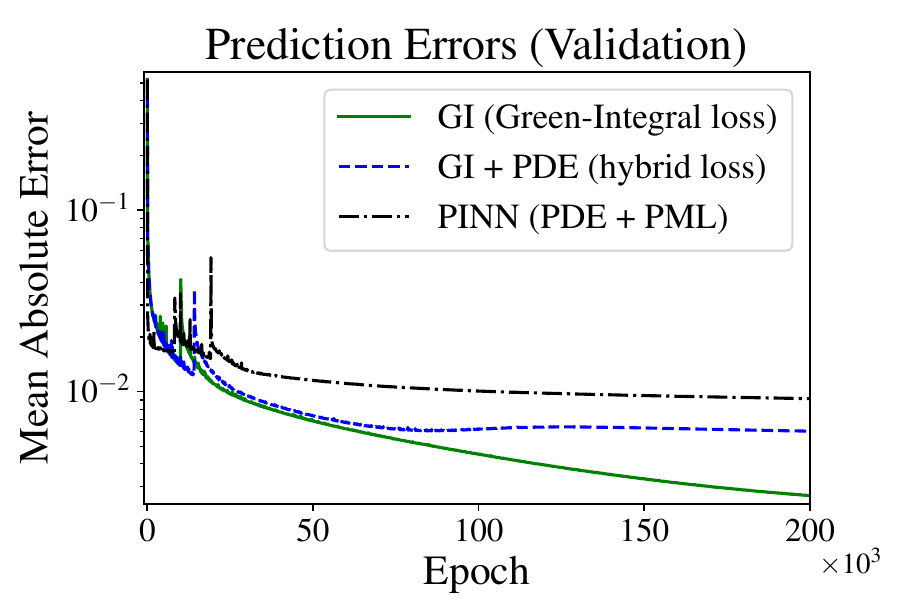}}
		\label{fig:otway_val_error}
	\end{subfigure}
	\\
	\begin{subfigure}{0.32\textwidth}
		\centering
		\stackinset{l}{-2pt}{t}{2pt}{\figlabel{d}}{%
			\includegraphics[width=\textwidth]{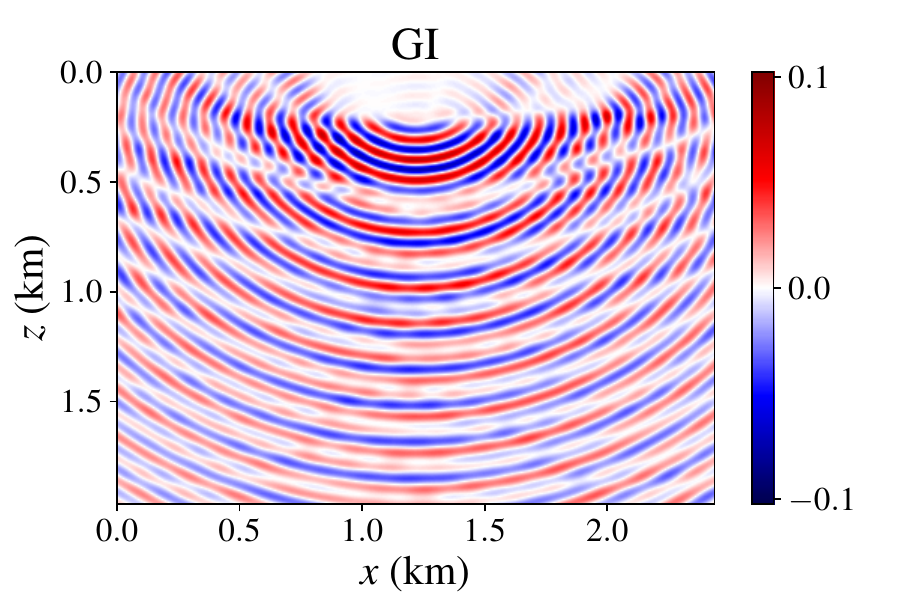}}
		\label{fig:otway_prediction_GI}
	\end{subfigure}
	\begin{subfigure}{0.32\textwidth}
		\centering
		\stackinset{l}{-2pt}{t}{2pt}{\figlabel{e}}{%
			\includegraphics[width=\textwidth]{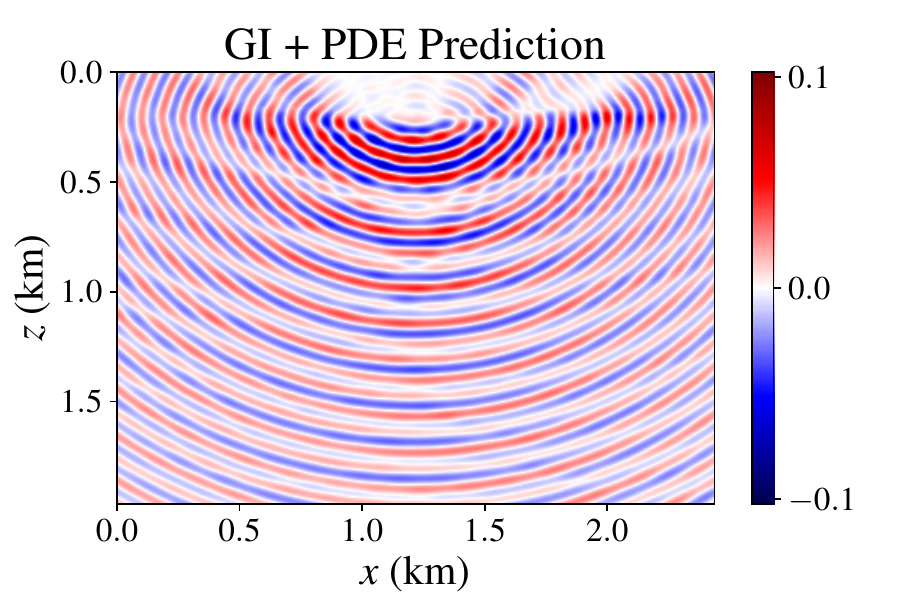}}
		\label{fig:otway_prediction_GIPDE}
	\end{subfigure}
	\begin{subfigure}{0.32\textwidth}
		\centering
		\stackinset{l}{-2pt}{t}{2pt}{\figlabel{f}}{%
			\includegraphics[width=\textwidth]{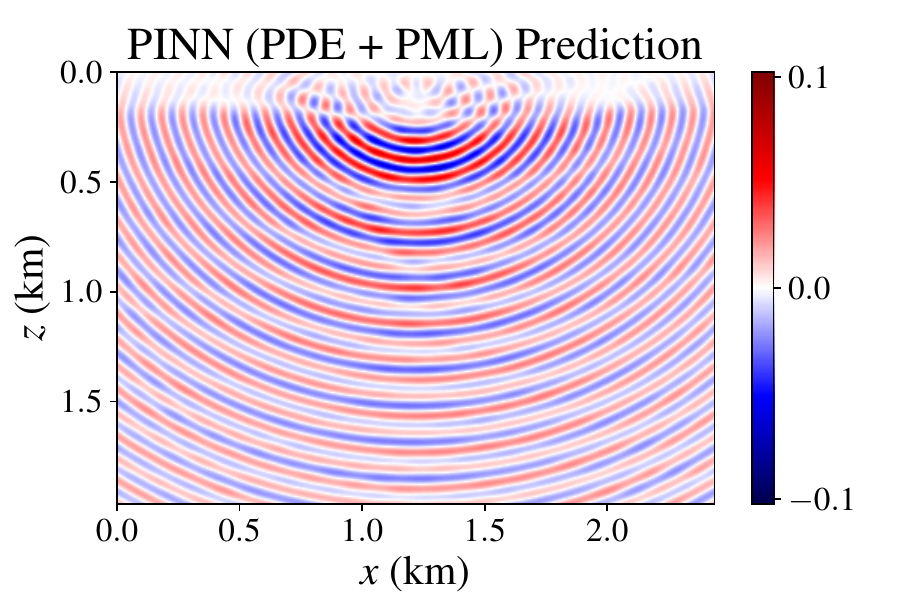}}
		\label{fig:otway_prediction_PDEPML}
	\end{subfigure}
	\\
	\begin{subfigure}{0.32\textwidth}
		\centering
		\stackinset{l}{-2pt}{t}{2pt}{\figlabel{g}}{%
			\includegraphics[width=\textwidth]{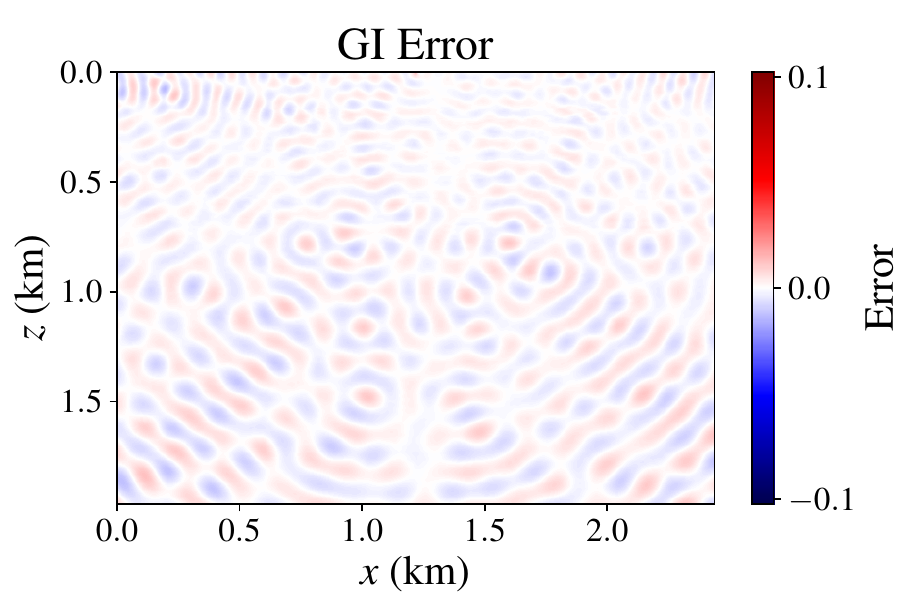}}
		\label{fig:otway_prediction_GI_error}
	\end{subfigure}
	\begin{subfigure}{0.32\textwidth}
		\centering
		\stackinset{l}{-2pt}{t}{2pt}{\figlabel{h}}{%
			\includegraphics[width=\textwidth]{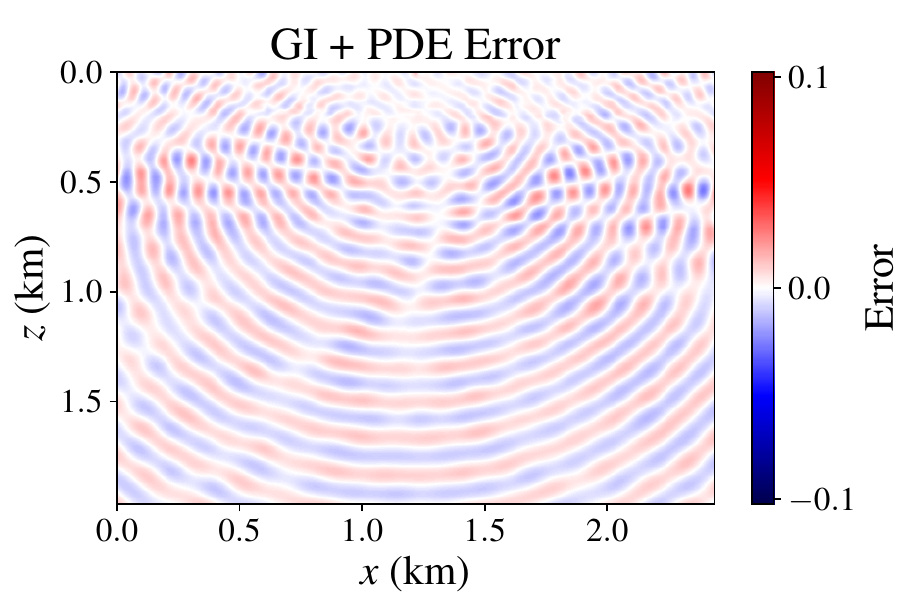}}
		\label{fig:otway_prediction_GIPDE_error}
	\end{subfigure}
	\begin{subfigure}{0.32\textwidth}
		\centering
		\stackinset{l}{-2pt}{t}{2pt}{\figlabel{i}}{%
			\includegraphics[width=\textwidth]{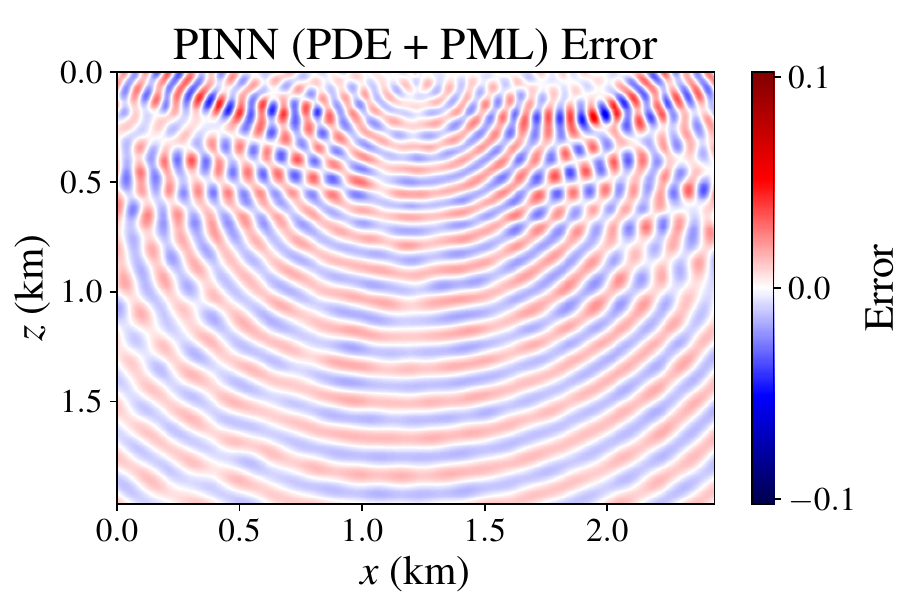}}
		\label{fig:otway_prediction_PDEPML_error}
	\end{subfigure}

\caption{
	Results for the Otway velocity model at 20~Hz.
	In the presence of dense subwavelength layering, the PINN formulation with PML fails to recover the correct scattering response, while GI-based training remains stable and accurate at lower computational cost (see Table~\ref{tab:otway_cost}).
	The hybrid GI+PDE loss degrades relative to GI-only training in this regime, indicating a limitation of local differential constraints for thin-layered media.
}
	\label{fig:otway_results}
\end{figure*}

We finally evaluate the proposed GI-based training on the Otway velocity model, a realistic geological model derived from the Otway Basin and representative of sedimentary environments with fine-scale stratification.
Unlike the Marmousi and Overthrust models, the Otway model is characterized by a large number of thin sedimentary layers whose thickness is smaller than the dominant wavelength (Figure~\ref{fig:otway_results}a).
For a 20~Hz wavefield, this results in wave propagation dominated by subwavelength heterogeneity, where the scattered wavefield is controlled by cumulative phase and amplitude effects rather than by isolated reflections from individual interfaces.

For reference, a 20~Hz scattered wavefield is simulated using FD modeling on a fine grid and presented on a grid of \(492 \times 610\) points for validation. The neural network architecture and optimization settings are kept identical to those used in the Overthrust experiment. For the GINN training, the GI loss is evaluated on a uniform grid of \(532 \times 330\) points (\(N_y = 175{,}560\)). For the PDE-based PINN training, using the same number of collocation points would demand 19~GB of GPU memory (by TensorFlow) and 17 hours of training; instead, a more practical number of \(35{,}000\) collocation points is employed to limit the computational costs.

Figure~\ref{fig:otway_results} summarizes the results for this test.
The first row shows the Otway velocity model, the FD reference scattered wavefield, and the evolution of the validation error during training.
The second row compares the predicted scattered wavefields obtained using GI-only training, hybrid GI+PDE training, and a PINN formulation based on the PDE residual with PML boundaries.

The PINN fails to converge to a sufficiently accurate solution for the Otway model. Although the PDE residual is minimized during training, the predicted wavefield resembles the response of a smoothed velocity model and does not capture the cumulative scattering effects induced by the fine layering.

The hybrid GI+PDE formulation initially follows the convergence trend of the GI-only model, but as $\lambda$ increases, it gradually diverges toward that of the PDE-based PINN. This behavior indicates that the local PDE constraint competes with the global integral consistency enforced by the GI term.

In comparison, GI-only training remains stable throughout the optimization, producing the most accurate reconstruction among the tested methods; even with increased frequency and model complexity, the GI formulation consistently captures the global scattering response.

A quantitative comparison of training cost and accuracy is reported in Table~\ref{tab:otway_cost}, which shows that GI-based training achieves substantially lower error at a significantly reduced computational cost compared to the PINN with PML.
These results highlight the robustness of the Green--Integral formulation in realistic velocity models dominated by fine-scale heterogeneity and motivate the discussion in the following section.

\begin{table}[]
	\centering
	\caption{Comparison of NN training methods for simulating the 20~Hz scattered wavefield in the Otway velocity model (see Figure~\ref{fig:otway_results}). GPU memory indicates peak usage as reported by TensorFlow.}
	\setlength{\tabcolsep}{3pt}
	\begin{tabular}{lccccc}
		\hline
		\textbf{Method} & $\mathbf{N_x}$ & $\mathbf{N_y}$ & \textbf{GPU Mem.} & \textbf{Time} & \textbf{NMSE} \\
		\hline
		\vspace{-5pt}\\
		GI (Green--Integral loss)      & 0        & 175{,}560 & \textbf{1.26} GB & \textbf{54} min  & $\textbf{0.024}$ \\
		GI + PDE (hybrid loss)         & 5{,}000  & 175{,}560 & 1.72 GB & 89 min  & $0.108$ \\
		PINN (PDE + PML)               & 35{,}000 & 0         & 3.83  GB& 214 min & $0.251$ \\
		\hline
	\end{tabular}
	\label{tab:otway_cost}
\end{table}

\section{Discussion}
\label{sec:discussion}
From the perspective of classical wave-equation solvers, the proposed Green--Integral (GI) neural network optimization can be interpreted as a learned preconditioned iterative method. As shown in Section~\ref{sec:connection_to_classical}, minimizing the GI residual corresponds to a Landweber-type iteration applied to the Lippmann--Schwinger system, while the neural parameterization introduces an implicit preconditioning operator related to the Neural Tangent Kernel. The spectral properties of this operator depend on the network architecture. From this viewpoint, the GI neural solver extends classical iterative approaches by introducing a learned, dynamically adapting preconditioner. 

In regimes of weak scattering where the Born--Neumann series converges, classical fixed-point iterations are the standard approach. However, in strongly heterogeneous media, the learned preconditioning provided by the neural network representation enables robust convergence.

%
%

\subsection{Error dynamics: PDE vs. integral formulations}
An important observation emerging from our numerical experiments is the different error distribution produced by PDE-based and GI-based training for the scattered Helmholtz problem. In the formulation considered here, the right-hand side of the scattered Helmholtz equation includes the product of the background wavefield and the scattering potential.
Regions with the background velocity (e.g., the water layer), where the scattering potential vanishes, satisfy a homogeneous Helmholtz equation. When training with a PDE-based loss, this region is only weakly constrained. As a result, during PINN training, the scattered wavefield in the background layer often exhibits larger errors at early and intermediate stages, and improves only after deeper regions are learned and information propagates indirectly through the optimization process.

In contrast, GI-based training explicitly enforces the global coupling between the scattered field and the spatial distribution of scatterers through the integral formulation. In regions where the scattering potential vanishes, the scattered wavefield is therefore not locally underdetermined, but is entirely induced by contributions from heterogeneities elsewhere in the model through the background Green's function. Consequently, the homogeneous background layer (e.g., the water layer) consistently emerges as one of the most accurate regions under GI training. Residual errors under the GI formulation instead tend to emerge from regions of strong scattering potential, where complex multiple-scattering effects and resolution limitations of the fixed integration grid pose the greatest challenge.

This distinction is observed in parts of the Otway velocity model, which is characterized by dense subwavelength layering. In this model, individual layers are thinner than the dominant wavelength, and the scattered wavefield is governed primarily by the cumulative phase and amplitude effects of many weak contrasts rather than by resolvable reflections from individual interfaces. While these cumulative effects are naturally constrained by the GI formulation, pointwise PDE residuals admit locally consistent solutions that correspond to an effective smooth or homogenized medium. As a result, PDE-based training (even when augmented with a perfectly matched layer) can converge toward homogenized wavefields that fail to capture the true scattering response.

The GI-based training effectively mitigates this low-frequency bias. Because oscillatory behavior and radiation characteristics are imposed directly through the integral kernel, smooth or overly homogenized solutions are strongly penalized early in training, and high-frequency phase and amplitude information is recovered more robustly. This explains both the sharper predictions of GI training in the Otway experiment and the observed degradation of hybrid GI+PDE training when local differential constraints begin to compete with global scattering consistency.

\subsection{Limitations and Future work}
The present study also has limitations. Efficient evaluation of the GI loss relies on a regular grid to enable the FFT-based computation, which may limit flexibility and local resolution in regions with sharp velocity contrasts or complex geometries. In addition, the present formulation assumes a homogeneous background Green's function and is restricted to acoustic wave propagation. The current implementation is also restricted to moderate frequencies representative of seismic-scale applications. Extending the proposed GI formulation to higher frequencies will require addressing increased resolution demands and computational complexity.

Despite these limitations, it is important to emphasize that the proposed Green--Integral framework is applicable to different neural network architectures and optimization algorithms. While the current study demonstrates its efficacy using a standard multilayer perceptron optimized via Adam, the GI loss can be readily integrated into other frameworks. For example, it could serve as an efficient physics-informed regularizer for neural operators, such as Fourier Neural Operators (FNOs) or Deep Operator Networks (DeepONets), or be applied to emerging architectures like Kolmogorov--Arnold, or Gabor-enhanced networks. Exploring these integrations to tackle higher-frequency regimes and elastic wave propagation represents a direction for future research.

\section{Conclusion}

We introduced a Green--Integral (GI) formulation for training neural networks to solve the scattered Helmholtz equation in heterogeneous media. By deriving a physics-based loss function, the proposed approach enforces global physical consistency between the predicted scattered wavefield and its integral reconstruction, fundamentally departing from conventional PDE-based PINN formulations. The GI loss uniquely constrains the solution, eliminates automatic differentiation for high-order spatial derivatives, enforces outgoing-wave behavior, and removes the reliance on perfectly matched layers.

Across a range of numerical experiments, including the Marmousi, Overthrust, and Otway velocity models, GI-based training demonstrated consistently improved stability, accuracy, and computational efficiency compared to PDE-residual-based PINNs. In particular, GI training achieved around an order-of-magnitude reduction in training time and GPU memory usage while producing more accurate wavefields. These gains become especially pronounced in settings where conventional PINN formulations suffer from severe optimization challenges.

We further investigated a hybrid GI+PDE training strategy, in which the global GI constraint is complemented by a local PDE residual evaluated at a small set of collocation points. This hybrid formulation showed  improve reconstruction accuracy in velocity models characterized by localized strong scatterers, where local corrections provide additional refinement beyond the fixed integration grid used for the GI loss. However, in velocity models dominated by dense subwavelength heterogeneity, such as the Otway model, the local PDE constraint was observed to compete against the global integral consistency, leading to degraded performance. These results highlight that the role of PDE residuals should be viewed as auxiliary corrections rather than a primary constraint.

Beyond performance improvements, the proposed framework provides physical insight into the limitations of local differential constraints for highly oscillatory wavefields. While minimizing the PDE residual enforces pointwise consistency with the governing operator, it admits homogenized or spurious solutions, particularly in subwavelength regimes. On the other hand, the nonlocal GI formulation preserves sensitivity to cumulative phase and amplitude effects induced by fine-scale heterogeneity.

Furthermore, this framework bridges deep learning with classical numerical linear algebra. We demonstrated that optimizing the GI loss via a neural network can mathematically be viewed as a spectrally preconditioned iteration. While the classical Born--Neumann series diverges in strongly scattering and high-contrast media, our method enables robust convergence using a suitable architecture.

While the present formulation is limited to acoustic wave propagation with a homogeneous background Green's function and moderate frequencies, the results demonstrate that enforcing global integral physics provides a robust alternative to PDE-residual-based training. Future work can extend the GI framework to higher frequencies, elastic and anisotropic media, three-dimensional domains, and inverse problems such as full waveform inversion.

Overall, the Green--Integral neural network framework offers a physically grounded, computationally efficient alternative to conventional PDE-based PINNs for wave propagation problems. By shifting the emphasis from local differential enforcement to global scattering consistency, it opens a promising pathway for scalable and robust neural solvers in complex heterogeneous media.

\section*{Acknowledgments}
The authors thank the DeepWave sponsors for their support. This research was supported by the following Research Projects/Grants: PID2023-146678OB-I00 funded by MICIU/AEI /10.13039/ 501100011033 and by the European Union Next Generation EU/ PRTR; “BCAM Severo Ochoa” accreditation of excellence CEX2021-001142-S funded by MICIU/AEI/ 10.13039/ 501100011033; Basque Government through the BERC 2022-2025 program; BEREZ-IA (KK-2023/00012) and RUL-ET(KK-2024/00086), funded by the Basque Government through ELKARTEK; Consolidated Research Group MATHMODE (IT1866-26) given by the Department of Education of the Basque Government; BCAM-IKUR-UPV/EHU, funded by the Basque Government IKUR Strategy and by the European Union Next Generation EU/PRTR. 

\section*{CRediT authorship contribution statement}
\textbf{Mohammad Mahdi Abedi:} Conceptualization, Methodology, Writing -- original draft, Data curation, Investigation, Software, Visualization.  
\textbf{David Pardo:} Conceptualization, Supervision, Writing -- review \& editing, Funding acquisition.  
\textbf{Tariq Alkhalifah:} Conceptualization, Supervision, Writing -- review \& editing, Funding acquisition. 

\section*{Declaration of competing interest}
The authors declare that they have no known competing financial interests or personal relationships that could have appeared to influence the work reported in this paper.

\section*{Code Availability}
A TensorFlow package that demonstrates the proposed implementation steps and instructions to reproduce the numerical examples is available at: \small{\\
\url{https://github.com/mahdiabedi/Green-Integral-Neural-Solver-for-the-Helmholtz-Equation}}

 \appendix
 \section{Cell-Averaged Self-Interaction of the 2D Helmholtz Green's Function}
 \label{app:self_term}
 \setcounter{equation}{0}
 \renewcommand{\theequation}{A-\arabic{equation}}
 In the numerical evaluation of the Green--integral equation \eqref{eq:green_discrete} for the 2D acoustic Helmholtz equation, the free-space Green's function becomes weakly singular. This appendix derives a cell-averaged self-interaction term to regularize the singularity while preserving compatibility with FFT-based convolution.
 
 Green's function for the 2D Helmholtz equation is given by
 \begin{equation}
 	G_0(\mathbf{x},\mathbf{y}) = \frac{i}{4} H_0^{(2)}(k_0 r),
 	\qquad r = \lVert \mathbf{x}-\mathbf{y} \rVert.
 \end{equation}
As $r \to 0$, the asymptotic expansion of the above equation yields
\begin{equation}
	G_0(r) =
	\frac{1}{2\pi} \ln r
	\;+\;
	\frac{1}{2\pi} \Big(\ln\frac{k_0}{2} + \gamma \Big)
	\;+\;
	\frac{i}{4}
	\;+\;
	\mathcal{O}(r^2),
	\label{eq:G0_asymptotic}
\end{equation}
 where $\gamma \approx 0.577$ is the Euler--Mascheroni constant. The first term is the logarithmic singularity, which is integrable, but it prevents pointwise evaluation at $r=0$.

 In equation \eqref{eq:green_integral}, the Green's function appears only under an integral. When $\mathbf{x}$ coincides with a discretization point $\mathbf{y}_k$, the corrected discrete contribution for $G_0(0)$ (which is undefined) is the \emph{cell-averaged value} of the Green's function over the finite grid cell associated with $\mathbf{y}_k$:
 \begin{equation}
 	G_0(0)
 	=
 	\frac{1}{W}
 	\int_{\text{cell}} G_0(\mathbf{r}) \, d\mathbf{r}.
 \end{equation}
Approximating the grid cell by a disk of equivalent area $W = \pi h^2$ with radius $h$, the cell-averaged logarithmic singularity is
 \begin{align}
 	\frac{1}{W} \int_{r<h}\frac{1}{2\pi} \ln r \, d\mathbf{r}
 	&= \frac{1}{\pi h^2} \int_0^h r \ln r \, dr \nonumber \\
 	&=\frac{1}{2\pi}\Big( \ln h - \frac{1}{2}\Big).
 \end{align}
 Adding the averaged singular contribution to the regular part from \eqref{eq:G0_asymptotic}, the final cell-averaged self-interaction term is
 \begin{equation}
 	G_0(0)
 	=
 	\frac{1}{2\pi} \left( \ln h + \ln\frac{k_0}{2} + \gamma - \frac{1}{2} \right)
 	+
 	\frac{i}{4}.
 	\label{eq:G_self_final_corrected}
 \end{equation}
The use of the Hankel function of first kind changes the sign of the real part of \eqref{eq:G_self_final_corrected}. This finite value replaces the undefined pointwise value $G_0(0)$ in the precomputed discretized Green's function kernel. It maintains translational invariance and is fully compatible with FFT-based convolution while avoiding the complexity of boundary-element techniques.

	\bibliographystyle{apalike}
\bibliography{references}
	
\end{document}